\documentclass[11pt]{article}

\usepackage[preprint]{acl}
\usepackage{times}
\usepackage{float}
\usepackage{CJKutf8}
\usepackage{latexsym}
\usepackage{multirow}
\usepackage{graphicx}
\usepackage{makecell}
\usepackage{amssymb}
\usepackage{pifont}
\usepackage{enumitem}
\usepackage{enumitem}
\usepackage{longtable}

\usepackage[T1]{fontenc}

\usepackage[utf8]{inputenc}

\usepackage{microtype}

\usepackage{float}
\restylefloat{table}
\usepackage{subcaption}
\usepackage{pdfpages}

\usepackage{inconsolata}

\usepackage{graphicx}
\usepackage{amsmath}
\usepackage{booktabs}
\usepackage{threeparttable}

\title{Lost in Translation: Investigating Systematic Discrepancies between Parallel English and Chinese Names of American Chinese Restaurants}

\author{
  \textbf{Hang Jiang\textsuperscript{1,4}},
  \textbf{Nanxi Liu\textsuperscript{1,2}},
  \textbf{Tiantian Wang\textsuperscript{4}},
  \textbf{Cassandra Overney\textsuperscript{1}},
\\
  \textbf{Rukun Zhang\textsuperscript{1}},
  \textbf{Artemisia Luk\textsuperscript{1}},
  \textbf{Diyi Yang\textsuperscript{3}},
 \textbf{Deb Roy\textsuperscript{1}},
\\
\textbf{Jad Kabbara\textsuperscript{1}}
\\
  \textsuperscript{1}Massachusetts Institute of Technology,
  \textsuperscript{2}Wellesley College,
  \textsuperscript{3}Stanford University,
  \textsuperscript{4}Northeastern University
\\
  \small{
    \textbf{Correspondence:} \href{mailto:email@domain}{hang.jiang@northeastern.edu}
  }
}

\begin{document}
\maketitle
\begin{abstract}
Given that the Chinese are one of the oldest ethnic groups to immigrate to the US, their cuisine have become central to the US dining scene and an interesting avenue to study how Chinese immigrants have culturally adapted into American life. In that regard, previous research has focused on analyzing the English names that Chinese restaurants adopt, e.g., finding the use of familiar and typical words. To investigate how Chinese immigrants have culturally integrated in the US depending on social, geographic and demographic contexts, we investigate in this work the under-explored relationship between the Chinese and English names of American Chinese restaurants. Specifically, we conduct a study on 3,172 Chinese restaurants (10\% of the total on Google Maps) sampled across the US. We present 8-dimension name frames to capture the underlying meanings behind names and create the \textsc{NameFrames} dataset containing 884 annotated parallel Chinese/English names. We also built an NLP pipeline to analyze the semantic discrepancies between names with both frame-based and embedding-based distance metrics. Our frame analysis shows that English names more often emphasize \textit{Location} and \textit{Specialty}, while Chinese names more often emphasize \textit{Positivity} and \textit{Ambiance}. Furthermore, we observe that for Chinese restaurants located in urban areas, those in counties with a higher concentration of Chinese population, or those that are relatively new, the discrepancies between the English and Chinese names are noticeably larger. Finally, interviews with 10 Chinese restaurant owners reveal additional interesting insights regarding naming conventions.

\end{abstract}

\section{Introduction}

\begin{figure}[h]
\includegraphics[width=\columnwidth]{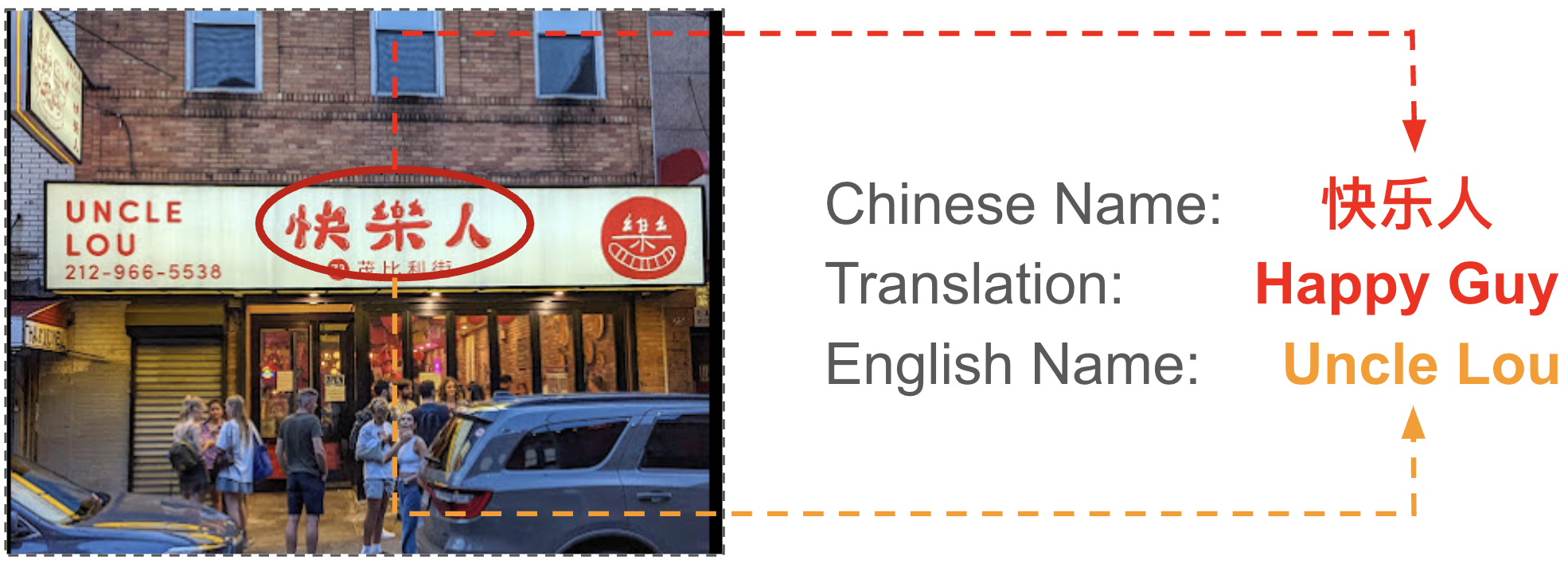}
\caption{An example of mismatched English and Chinese ACR names. We hired Prolific workers to transcribe the names by browsing photos on Google Maps. The English name encodes the \textit{Personal Name} frame, while the Chinese name encodes the \textit{Positivity} frame.}
\label{fig:chinese_name_transcription}
\end{figure}

The process of immigrant groups adapting their language, values, and behaviors to a new society is commonly known as \textbf{cultural assimilation} in sociology and ethnic studies. The United States, colloquially hailed as ``a melting pot'', has a long history of immigration and therefore provides a unique perspective to researchers to study this phenomenon. Among the diverse immigrant groups in the US, those of Chinese ethnicity (referred to as ``Chinese'') hold a significant place, because their path to assimilation is not in a ``straight-line'' fashion \cite{warner1945social,alba2014rethinking} but somehow ``bumpy'' and ``segmented'' \cite{portes1993new,zhou2005multifaceted}. Over time, Chinese immigrants have established themselves through the restaurant business, which is central to the dining scene in the United States. Different from other ethnic restaurants in the US, Chinese restaurants usually have parallel names in English and Chinese, which sometimes do not match in meaning (Figure \ref{fig:chinese_name_transcription}). These parallel names reflect both an embrace of cultural assimilation and a preservation of their home culture identity to Chinese immigrants. \textbf{This paper provides a systematic analysis of this cultural phenomenon by investigating patterns of divergence between Chinese and English names for over three thousand American Chinese restaurants (ACRs).}

Prior work has used ACR menus and reviews \cite{chen2021analyzing,luo2023othering} to study cultural assimilation of the Chinese community. This work focuses on names because names represent how people define and mark their own cultural identity and we are interested in studying how ACR owners represent themselves as immigrants from their ACR naming practice. \citet{ferdman2016we} and \citet{chin2023symbols} have observed that ACR English names frequently feature cultural symbols such as ``panda'', ``dragon'', and ``wok''. Through a qualitative analysis on 423 ACRs in Los Angeles County, \citet{chen2018authenticity} further categorized ACR names into various frames such as \textit{Place Names}. This study revealed how ACR owners demonstrate cultural assimilation through the adoption of English names, terms evoking modern European sensibility, and concepts related to American fast-food takeout. However, previous research has never compared the Chinese names with the accompanying English names in ACRs.

This paper presents a novel study on the cultural assimilation of Chinese in American society by focusing, as a case study, on the phenomenon of systematic discrepancies between parallel English and Chinese names in ACRs. We sampled 3,172 American Chinese restaurants (ACRs), 10\% of the total on Google Maps, across the US and enrolled crowdworkers (67 hours in total) to transcribe their Chinese names by reviewing Google Map photos and websites. Building on \citet{chen2018authenticity}'s work, we adapt 8-dimension name frames to capture various aspects of the underlying meaning of names. To automate the analysis, we build an NLP pipeline to compare the discrepancies between the parallel names with both frame-based and embedding-based distance metrics. Finally, we examine how the difference in names is correlated with socio-contextual factors, contributing significantly to our understanding of this unique cultural practice. Our main contributions are as follows:
\begin{itemize}
    \item We present a novel study that examines the cultural assimilation of Chinese in the US and uncovers systematic semantic discrepancies between English and Chinese names of ACRs. 
    \item We adapt 8-dimension name frames based on prior work to capture underlying meanings of names and create the \textsc{NameFrames} dataset on 884 pairs of ACR names.
    \item We build an NLP pipeline to quantify semantic discrepancies with two  metrics, showing that for ACRs in urban areas, those in counties with a higher percent of Chinese, or those that are relatively new, the discrepancies between their parallel names are significantly larger.
    \item We conduct ten qualitative interviews with ACR owners to further complement the computational analysis, which helps us better understand their motivations and experiences behind their naming practices.
\end{itemize}

\section{Related Work}

\subsection{Names of American Chinese Restaurants}

Traditionally, running American Chinese restaurants (ACRs) is an important means of livelihood for Chinese immigrants in the United States, leading to the birth of what we now know as American Chinese cuisine \cite{coe2009chop}. For instance, the American national condiment ``ketchup'' is linked to the history of Fujianese immigrants opening Chinese restaurants in New York City 
\cite{jurafsky2014language}. Researchers from Washington Post \cite{ferdman2016we} and Smithsonians\footnote{\url{https://www.smithsonianmag.com/smithsonian-institution/why-do-chinese-restaurants-such-similar-names-\\180961996/}} noticed that the English names of ACRs tend to be similar and stereotypical, such as ``wok'', ``garden'',  ``dragon'', etc. \citet{chen2018authenticity} did a qualitative study on the English naming practices of 423 ACRs in Los Angeles County, demonstrating how these names, through the incorporation of various name frames such as region names, family names, and transliterated hanzi, reflect the diversity within Chinese culinary traditions and the immigrants' adaptation to American culture. Despite these insightful contributions, none of the previous work delved into the patterns existing between parallel English and Chinese names of ACRs. Previous studies have predominantly focused on the English names of these restaurants \cite{liu2009food,ferdman2016we,chen2018authenticity,li2019corpus}, often overlooking the rich cultural narratives and adaptive strategies embedded in the parallel names. The discrepancies between these parallel names can offer a deeper understanding of the cultural adaptation processes of Chinese immigrants and their attempts to negotiate identity and representation in a multicultural context.

\subsection{Computational Frame Analysis}
Prior work in NLP has presented various computational methods to automate framing analysis in large-scale datasets \cite{ziems-yang-2021-protect-serve,ali-hassan-2022-survey,ziems2023can}. If there is no pre-annotated dataset or codebook, clustering algorithms \cite{burscher2016frames} and topic modeling \cite{dimaggio2013exploiting,nguyen2015guided} are two popular unsupervised approaches to analyze frames on large corpora. However, supervised neural networks are commonly applied to detect frames if there is an existing coding schema with enough annotated frame data for training \cite{ali-hassan-2022-survey}, including  probabilistic soft logic \cite{johnson-etal-2017-leveraging}, LSTM \cite{naderi2017classifying}, recursive neural network \cite{ji2017neural}, and transformer-based language models such as BERT \cite{liu-etal-2019-detecting,tourni-etal-2021-detecting-frames}, RoBERTa \cite{khanehzar-etal-2019-modeling,huguet-cabot-etal-2020-pragmatics,mendelsohn-etal-2021-modeling}, XLM-RoBERTa \cite{bhatia-etal-2021-openframing}, and  MultiBERT \cite{akyurek-etal-2020-multi}. Recently, large language models have shown promising results to automate social science research \cite{grossmann2023ai,bail2023can,rytting2023towards,shen2023shaping,karjus2023machine}, including automatic frame detection and analysis \cite{ziems2023can,tornberg2023chatgpt} in a zero-shot or few-shot setting. 

\subsection{Measuring Semantic Textual Similarity}
Estimating the semantic textual similarity (STS) between text data is an open problem in NLP \cite{chandrasekaran2021evolution}. In earlier approaches, two text snippets were considered similar if they contain the same words or n-grams. However, lexicon-matching methods such as BoW and TF-IDF fail to consider that different words can share similar meanings. To address the drawback of the lexical measures, prior work has proposed knowledge-based methods such as edge-counting \cite{rada1989development} and information content (IC)-based \cite{sanchez2012ontology,zhu2016computing} approaches, corpus-based methods such as LSA \cite{landauer1997solution} and LDA \cite{sinoara2019knowledge}, and neural networks-based methods such as word2vec \cite{wang-etal-2016-sentence,hamilton-etal-2016-diachronic,hamilton-etal-2016-cultural,kutuzov-etal-2018-diachronic}, LSTM \cite{tai-etal-2015-improved}, and CNN \cite{he-lin-2016-pairwise} to measure textual similarity or quantify semantic change. Recently, transformer-based models such as BERT \cite{devlin-etal-2019-bert}, RoBERTa \cite{liu2019roberta}, DistilBERT \cite{sanh2019distilbert}, and T5 \cite{raffel2020exploring} are pre-trained as language models and have been commmonly used to represent texts for measuring textual similarity. \citet{reimers-gurevych-2019-sentence} introduced Sentence-BERT (SBERT), which can derive semantically meaningful sentence embeddings that can be compared using cosine similarity. \citet{reimers-gurevych-2020-making} extended SBERT to support multiple languages. The most recent work in NLP focus on optimizing language models for STS \cite{gao-etal-2021-simcse,chuang-etal-2022-diffcse,wu-etal-2022-esimcse,jiang-etal-2022-promptbert,li2023angle}, improved smaller language models by generating synthetic data from LLMs \cite{wang2023improving}, or scaling sentence embeddings to larger language models such as T5 \cite{ni2021sentence} and 2.7B OPT \cite{jiang2023scaling}.


\section{\textsc{NameFrames} Dataset}

We collected all American Chinese Restaurants (ACRs) with Google Maps API, sampled a subset of ACRs from each state across the U.S., and hired crowdworkers to double-transcribe Chinese names from online images. To quantify the information lost in translation, we adapted 8-dimensional name frames and created the \textsc{NameFrames} dataset. This dataset includes annotations of both English and Chinese names across 884 ACRs. 

\subsection{Data Collection}
From 2023-07-13 to 2023-07-15, we collected U.S. restaurants categorized as ``Chinese restaurant'' using Outscrapper\footnote{\url{https://outscraper.com/google-maps-api/}}, which provides public Google Maps Places API information such as names, websites, locations, ratings, prices, and photos. We filtered out chain and fast food restaurants because their names repeat frequently (Appendix \ref{appendix:acr_filtering}), yielding 31,548 unique ACRs. We then used stratified sampling by state to select 3,172 ACRs: 10\% from states with more than 400 ACRs, 40 restaurants from states with 40--400 ACRs, and all restaurants from states with fewer than 40 ACRs. Finally, we removed duplicate Chinese--English name pairs. See Appendix \ref{appendix:acr_stats} for the distribution. See the distribution of ACRs in Appendix \ref{appendix:acr_stats}.




\setlength{\extrarowheight}{3pt}
\begin{table*}[ht!]
    \centering
    \footnotesize
    \begin{tabular}{ccc}
        {\textbf{Frame}}  & 
        {\textbf{Simplified Definition}}  &
        {\textbf{Selected Examples}} \\
         \Xhline{3\arrayrulewidth}
        \textit{Personal Name} & first or last names with or without possessives & \textit{\textbf{Lin’s} Kitchen}, \textit{\textbf{P.F. Chang's}} \\
        \textit{Specialty} & special dishes, cooking methods, or ingredients & \textit{\textbf{Peking Duck} House}, \textit{\textbf{Noodle} Galaxy} \\
        \textit{Positivity} & superlatives or other positive modifiers & \textit{\textbf{Happy} Lamb}, \textit{\textbf{Best} Chinese Buffet} \\
        \textit{Culture} & references to cultural associations & \textit{\textbf{Panda} Express}, \textit{The \textbf{Phoenix} Club} \\
        \textit{Location} & place names or regional origins & \textit{\textbf{China} Garden}, \textit{\textbf{Oriental} Kitchen}\\
        \textit{Ambiance} & dining atmosphere or intimate words of family or relationship & \textit{Jade \textbf{Garden}}, \textit{\textbf{Friendship} BBQ} \\
        \textit{Creativity} & puns or names derived from creative linguistic processes & \textit{\textbf{Wok N' Roll}}, \textit{\textbf{Noodology}} \\
        \textit{Romanization} & Romanized Chinese terms excluding locations and personal names & \textit{\textbf{Ding Tai Fung}}, \textit{\textbf{Hai Di Lao} Hot Pot} \\
        \Xhline{3\arrayrulewidth}
    \end{tabular}
    \caption{This table shows 8 frames with their simplified definitions and selected examples. The bold text spans indicate where the corresponding frame is encoded in the names. ``Ding Tai Fung'' means ``wok, peace, and abundance''. ``Hai Di Lao'' means ``catch from the deep ocean''. See full details of the frames in Appendix \ref{appendix:codebook}.}
        \label{tab:frame_definition}
\end{table*}

\subsection{Chinese Name Transcription}
Google Maps API provides English names of the ACRs but not the Chinese names. To obtain accurate transcriptions of their Chinese names, we first obtained IRB exemption and hired two Chinese-speaking crowd workers to identify and transcribe the Chinese name of each restaurant in simplified Chinese characters by browsing its photos from Google Maps (see Figure \ref{fig:chinese_name_transcription}). If there is any disagreement between two annotators, one of the authors reviewed the annotations to make a final decision. We observed a decent inter-rater agreement score 0.673 in Cohen's Kappa. After transcribing the Chinese names of the 3,172 ACRs, we identified 1,722 ACRs with both English and Chinese names and 1,450 with only English names.

\subsection{Name Frame Design}
\citet{chen2018authenticity} proposed a few name frames to conduct qualitative analysis on meanings encoded in ACR English names. We adapted their framework and generalized these frames to both English and Chinese names. Table \ref{tab:frame_definition} shows the simplified frame definitions and selected examples. The complete codebook 
can be found in Appendix \ref{appendix:codebook}.

\subsection{Name Frame Annotation}
We sampled 884 ACRs for frame annotation. Each restaurant was annotated by three co-authors who are native Chinese speakers with professional English proficiency. We annotated the data in two rounds, discussing disagreements after each round to refine the frame definitions and annotations. As shown in Table \ref{tab:annotation_agreement}, inter-annotator agreement was high across the eight frames, with scores ranging from 0.78 to 0.935 for English names and 0.79 to 0.9 for Chinese names. To tune LLMs for labeling the remaining data, we randomly split the annotated data into 50\% validation and 50\% test sets for each language, removing duplicate names within each language.




\section{Methods}

First, we describe methods to detect name frames and compare their performance on \texttt{NameFrames}. These frame detection models were used to automate frame analysis and to determine the information gained or lost when comparing English and Chinese names. We utilize both frame-based Jaccard distance and embedding-based Cosine distance to quantify the semantic discrepancies between names. These distance scores were subsequently correlated with socio-contextual factors. Finally, we conducted and analyzed ten qualitative interviews to deepen our understanding of the Chinese immigrant integration into US society.


\setlength{\extrarowheight}{3pt}
\begin{table*}[ht!]
    \centering
    \footnotesize
    \resizebox{\textwidth}{!}{%
\begin{tabular}{p{2.5cm}ccccccccc}
        {\textbf{Model}}  & 
        {\textbf{Personal Name}}  &
        {\textbf{Speciality}} & 
        {\textbf{Positivity}} & 
        {\textbf{Culture}} &
        {\textbf{Location}} & 
        {\textbf{Ambiance}}& 
        {\textbf{Creativity}}& 
        {\textbf{Romanization}} & 
        {\textbf{Average}}\\
         \Xhline{3\arrayrulewidth}
        \multicolumn{10}{c}{\textit{\textbf{Chinese Names} (Weighted F1)}} \\ 
        Expert Rules & $88.09$ & $97.39$ & $90.81$ & $94.79$ & $88.18$ & $93.44$ & $95.20$ & $-$ & $92.56$ \\
        Qwen3.7-Max & $95.14$ & $\mathbf{98.86}$ & $92.38$ & $94.15$ & $92.80$ & $\mathbf{96.08}$ & $96.89$ & $-$ & $95.18$ \\
        Kimi K2.6 & $94.23$ & $97.42$ & $88.64$ & $94.93$ & $94.37$ & $93.46$ & $\mathbf{97.05}$ & $-$ & $94.30$ \\
        GPT-5.5 & $\mathbf{96.01}$ & $98.02$ & $\mathbf{92.68}$ & $95.38$ & $93.99$ & $94.71$ & $96.88$ & $-$ & $\mathbf{95.38}$ \\
        Claude Sonnet 4.6 & $94.83$ & $97.75$ & $90.13$ & $\mathbf{96.66}$ & $\mathbf{95.20}$ & $94.95$ & $96.88$ & $-$ & $95.20$ \\
        \Xhline{3\arrayrulewidth}
        \multicolumn{10}{c}{\textit{\textbf{English Names} (Weighted F1)}} \\
        Expert Rules & $85.83$ & $85.83$ & $82.46$ & $93.95$ & $92.02$ & $\mathbf{91.43}$ & $74.22$ & $72.42$ & $84.77$ \\
        Qwen3.7-Max & $96.79$ & $\mathbf{97.46}$ & $91.61$ & $96.95$ & $91.11$ & $90.34$ & $\mathbf{99.21}$ & $91.79$ & $94.41$ \\
        Kimi K2.6 & $96.27$ & $96.72$ & $89.41$ & $\mathbf{97.64}$ & $91.61$ & $88.44$ & $97.60$ & $90.57$ & $93.53$ \\
        GPT-5.5 & $\mathbf{98.45}$ & $97.27$ & $\mathbf{93.14}$ & $97.49$ & $\mathbf{95.18}$ & $87.04$ & $97.01$ & $93.12$ & $\mathbf{94.84}$ \\
        Claude Sonnet 4.6 & $97.62$ & $97.27$ & $92.46$ & $96.35$ & $86.46$ & $85.95$ & $97.98$ & $\mathbf{94.08}$ & $93.52$ \\
        \Xhline{3\arrayrulewidth}
\end{tabular}
    }
    \caption{Performance of different approaches on Name Frame Detection in both English and Chinese ACR names. ``Expert Rules'' represents the rule-based system based on lexicons and heuristics. Qwen3.7-Max, Kimi K2.6, GPT-5.5, and Claude Sonnet 4.6 represent the best performance of each model across two prompt settings: (1) zero-shot, (2) few-shot, as described in Section \ref{sec:nfd_task}.}
        \label{tab:nfd_performance}
\end{table*}

\subsection{Name Frame Detection}
\label{sec:nfd_task}
\paragraph{Task Formulation} The task of name frame detection is defined in the following manner: Given a specific name denoted as $N$, our target is to identify if each specific frame $f_i$ within the set $F$ of all possible frames is encoded in the name $N$. 



\paragraph{Methods} We built rule-based and prompt-based methods to detect name frames from ACR English and Chinese names. To establish a strong baseline, we first developed a rule-based model (``Expert Rules'' in Table \ref{tab:nfd_performance}) based on lexicons and/or heuristics for each name frame respectively. Details about the lexicon-based approach are discussed in Appendix \ref{appendix:rule_model}. Afterwards, we experimented with GPT-5.5\footnote{\url{https://openai.com/index/introducing-gpt-5-5/}}, Qianwen (\texttt{qwen3.7-max})\footnote{\url{https://openrouter.ai/qwen/qwen3.7-max}}, Kimi (\texttt{kimi-k2.6})\footnote{\url{https://huggingface.co/moonshotai/Kimi-K2.6}}, and Claude Sonnet 4.6 (\texttt{claude-sonnet-4.6}) \footnote{\url{https://openrouter.ai/anthropic/claude-sonnet-4.6}} given that they are among the state-of-the-art multilingual LLMs. We do not include fine-tuned language models such as RoBERTa \cite{liu2019roberta} as baselines because they do not perform well on our small dataset where some labels are too sparse for effective fine-tuning.

\paragraph{Prompts} We experimented with three prompting strategies: (1) name + definition (``zero-shot''), (2) name + definition + rules + examples (``few-shot''). The first strategy is \textbf{zero-shot}, which prompts LLMs to predict if the restaurant name encodes a certain frame given its name and definition. The second strategy \textbf{few-shot}, which aims to improve the strategy (1) by providing a few examples or rules of the target frame along with its name and definition. Strategy (2) provides the frame name, definition, expert-curated rules, and common types of examples to LLMs in a highly similar format as the codebook (see Appendix \ref{appendix:codebook}). The format of these two prompt variants are discussed in the Appendix \ref{appendix:llm_model}.


\paragraph{Model Settings} For model-based name-frame labeling, we used comparable settings across GPT-5.5, Qwen3.7-Max, Kimi K2.6, and Claude Sonnet 4.6 where supported: seed 42, reasoning disabled, batches of 20 labels per request, and an 800-token output cap. We set temperature to 0.0 for Qwen3.7-Max, Kimi K2.6, and Claude Sonnet 4.6; GPT-5.5 did not support user-specified temperature for this endpoint, so we relied on seed 42 for determinism.


\paragraph{Results} In Table \ref{tab:nfd_performance}, we show the best results from each model due to space limits, with full prompt-setting results in Appendix \ref{appendix:full_nfd_performance}. Overall, no single model dominates across all frames, but the newer LLMs generally outperform Expert Rules. GPT-5.5 achieves the highest average weighted F1 in both Chinese and English labeling, so we use it to label the remaining data.

\subsection{Measuring Name Discrepancies}
We developed two different metrics to quantify discrepancies between the Chinese and English names. The first metric is based on name frames, which aims to capture typical meanings such as \textit{culture} and interesting syntactic forms such as \textit{creativity}. The second embedding-based metric is a standard NLP method to quantify semantic differences. In our study, we first translates the Chinese name to English and measures the embedding distance between translated and original English names. These two metrics are meant to capture slightly different aspects of linguistic discrepancies in names. 


\subsubsection{Frame-based Jaccard Distance (JD)}
Name frames encode nuanced linguistic differences in meanings (e.g., \textit{culture}) and forms (e.g., \textit{creativity}) between English and Chinese names. Therefore, we can use frame-based Jaccard distance ($JD$) to measure the discrepancies between English and Chinese names. We define $JD$ to be: $$JD(F_{EN}, F_{CH})=1-\frac{|F_{EN} \cap F_{CH}|}{|F_{EN} \cup F_{CH}|}$$



In the equations above, $F_{EN}$ and $F_{CH}$ indicates the set of name frames encoded in the English and Chinese names of the same restaurant respectively.




\subsubsection{Embedding-based Cosine Distance (CD)}

Alternatively, we measure semantic difference using name embeddings. We translate each Chinese restaurant name into English as literally as possible, then compute the distance between the translated Chinese name and the observed English name. Because literal name translation differs from standard machine translation, we evaluated Google Cloud Translation API Basic V2\footnote{\url{https://cloud.google.com/translate}}, GPT-5.5, and Claude Sonnet 4.6 against two human literal-translation references for 99 restaurants. Claude Sonnet 4.6 achieved the strongest average embedding similarity and BERTScore, so we use its translations in the main analysis; full translation results are reported in Appendix \ref{appendix:gpt_translation_performance}. We then use \texttt{all-mpnet-base-v2}\footnote{\url{https://huggingface.co/sentence-transformers/all-mpnet-base-v2}} to compute embedding-based semantic distance.


$$CD(E_{TS}, E_{EN})=1 - \frac{E_{TS} \cdot E_{EN}}{\|E_{TS}\|_2 \cdot \|E_{EN}\|_2}$$

In the equation above, $E_{TS}$ and $E_{EN}$ indicate the embeddings of the translated and original English names of the same restaurant respectively.

\subsubsection{Interview with ACR Owners}

To understand how ACR owners create Mandarin and English names, we interviewed ten ACR owners in Massachusetts about factors influencing naming decisions like customer groups and location, and how the restaurant reflects the owner's heritage and identity. We analyzed the interviews using affinity diagramming~\cite{lucero2015using} to identify key naming themes. Additional interview procedures are in Appendix~\ref{app:interviewProcedure}, and a fuller summary of results is in Appendix~\ref{app:interviewResults}.

\section{Analysis \& Discussion}
\label{sec:analysis}

We are interested in answering the following research questions (RQs): \textbf{RQ1}: are there systematic discrepancies in naming practices in ACRs? \textbf{RQ2}: what socio-contextual factors impact the ACR naming process? \textbf{RQ3}: do ACR owners deliberately adjust the parallel names to their audiences? 




\subsection{Discrepancies in Parallel Names}
\begin{figure}[h]
\includegraphics[width=\columnwidth]{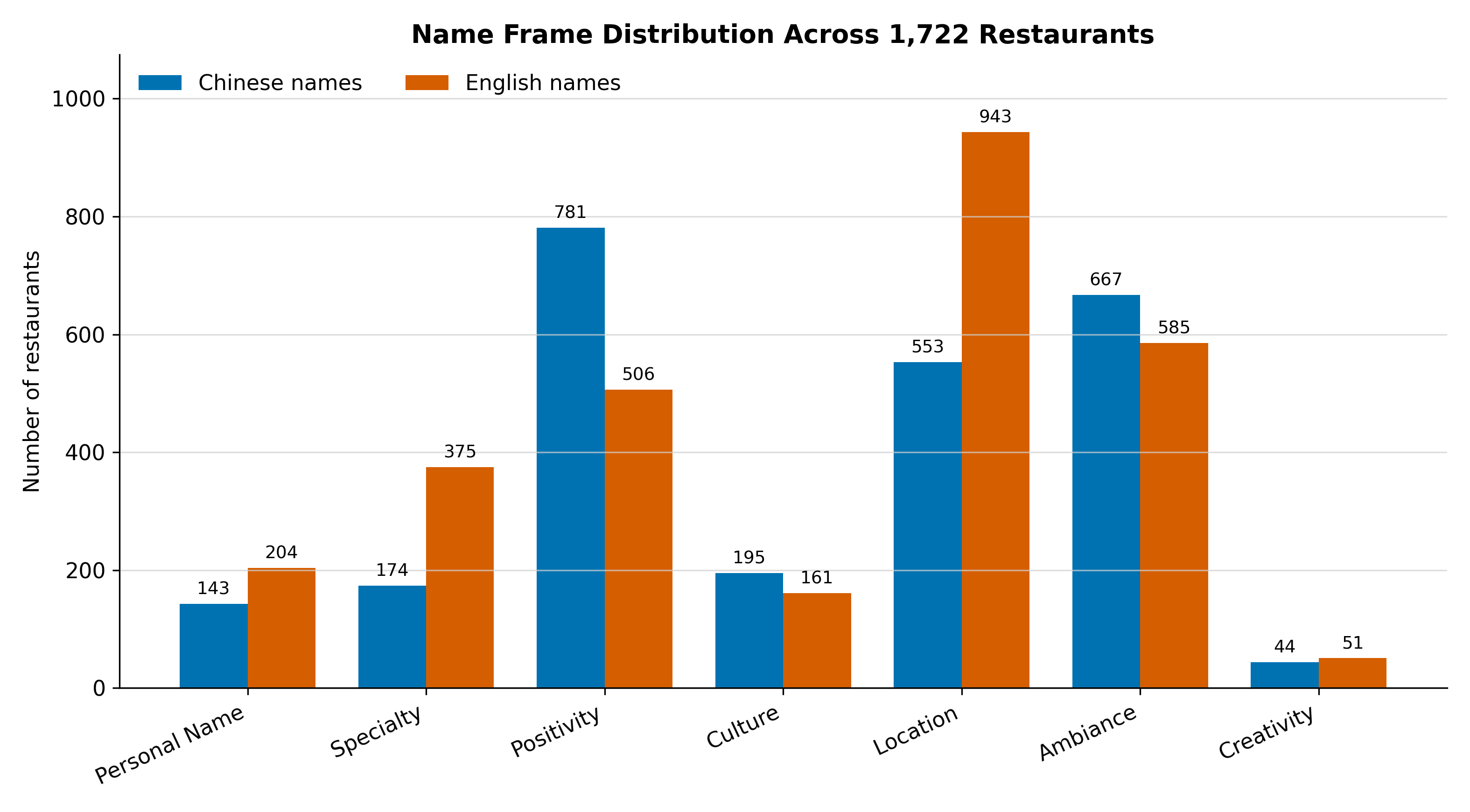}
\caption{Percentage of unique English and Chinese names tagged with each frame.}
\label{fig:qwen_frame_dist}
\end{figure}

By applying the best frame detection models to the 1,722 ACRs with both Chinese and English names, we show the distribution of naming frames in Figure \ref{fig:qwen_frame_dist}, using GPT5.5 labels with human overrides when available. The clearest pattern is that English names more often emphasize \textit{Location} and \textit{Specialty}, while Chinese names more often emphasize \textit{Positivity}. Chinese names also show slightly higher counts for \textit{Culture} and \textit{Ambiance}, while \textit{Creativity} remains rare in both languages. These patterns suggest that English-facing names tend to foreground legible market information, such as geographic cues and cuisine type, whereas Chinese names more often preserve auspicious, evaluative, or culturally meaningful content.

\subsection{Location}

\begin{figure*}
    \centering
    \begin{subfigure}[b]{0.3\textwidth}
        \includegraphics[width=\textwidth]{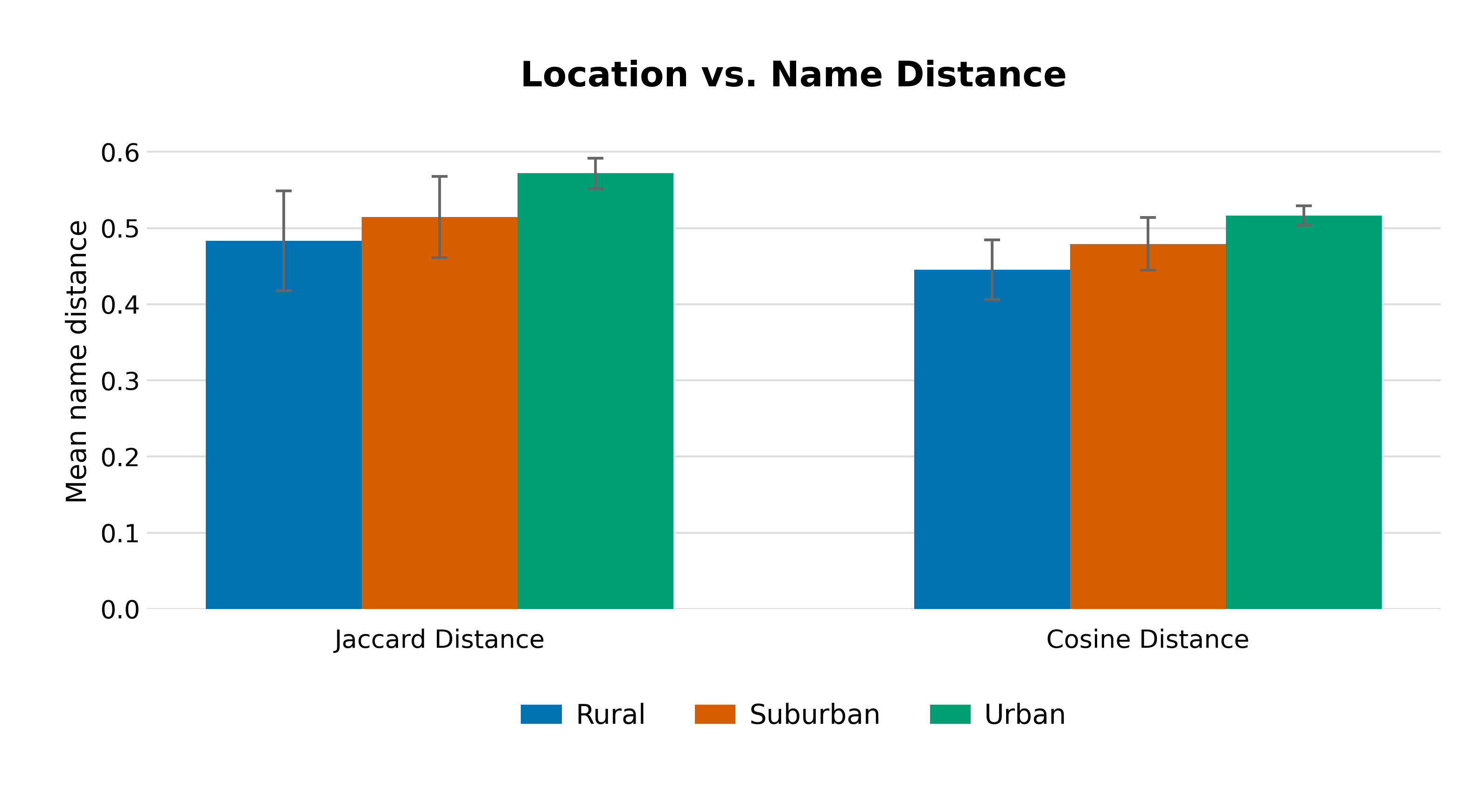}
        \caption{Location}
        \label{fig:location_distance}
    \end{subfigure}%
    ~ 
    \begin{subfigure}[b]{0.3\textwidth}
        \includegraphics[width=\textwidth]{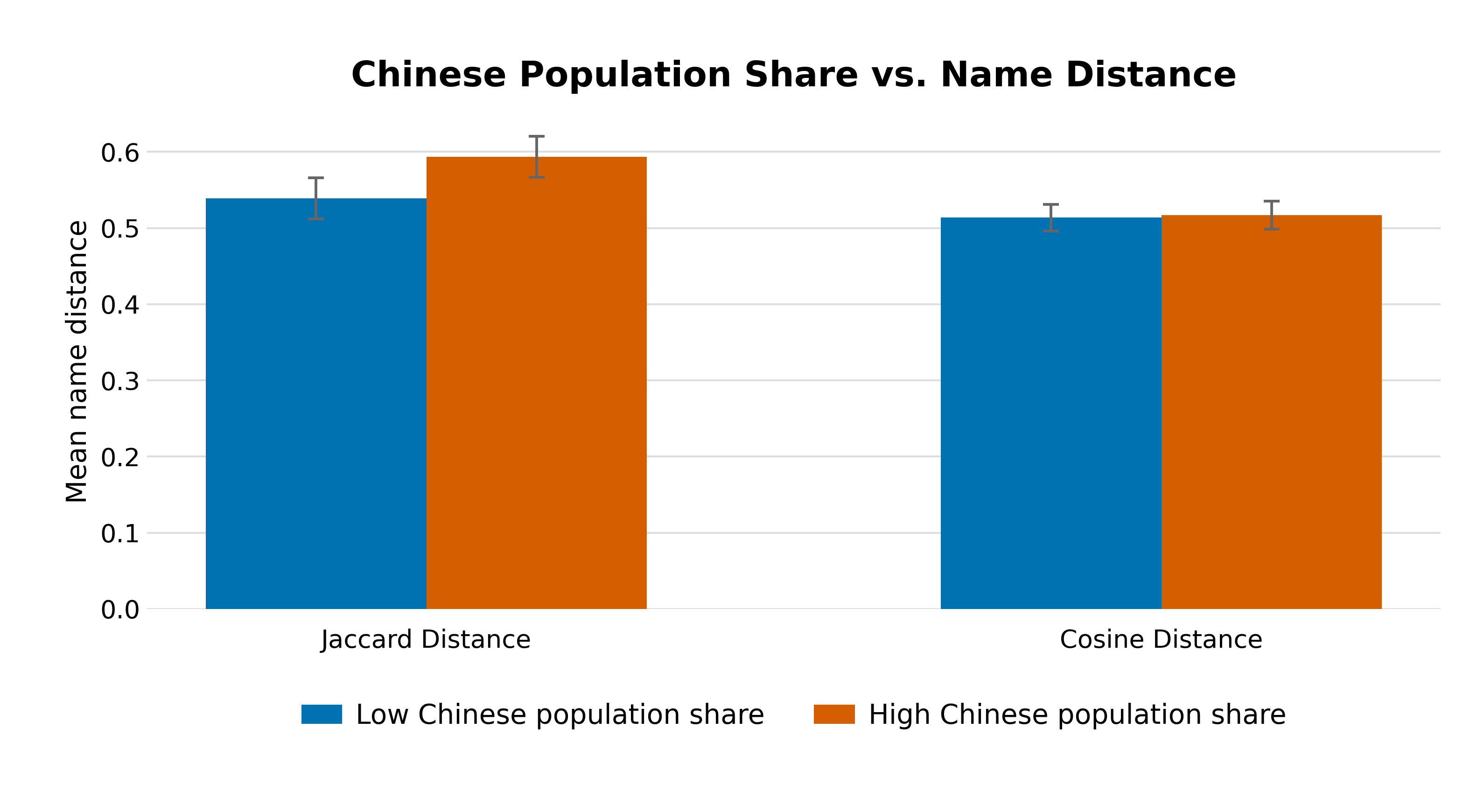}
        \caption{Chinese Percentage}
        \label{fig:chinese_distance}
    \end{subfigure}
    ~ 
    \begin{subfigure}[b]{0.3\textwidth}
        \includegraphics[width=\textwidth]{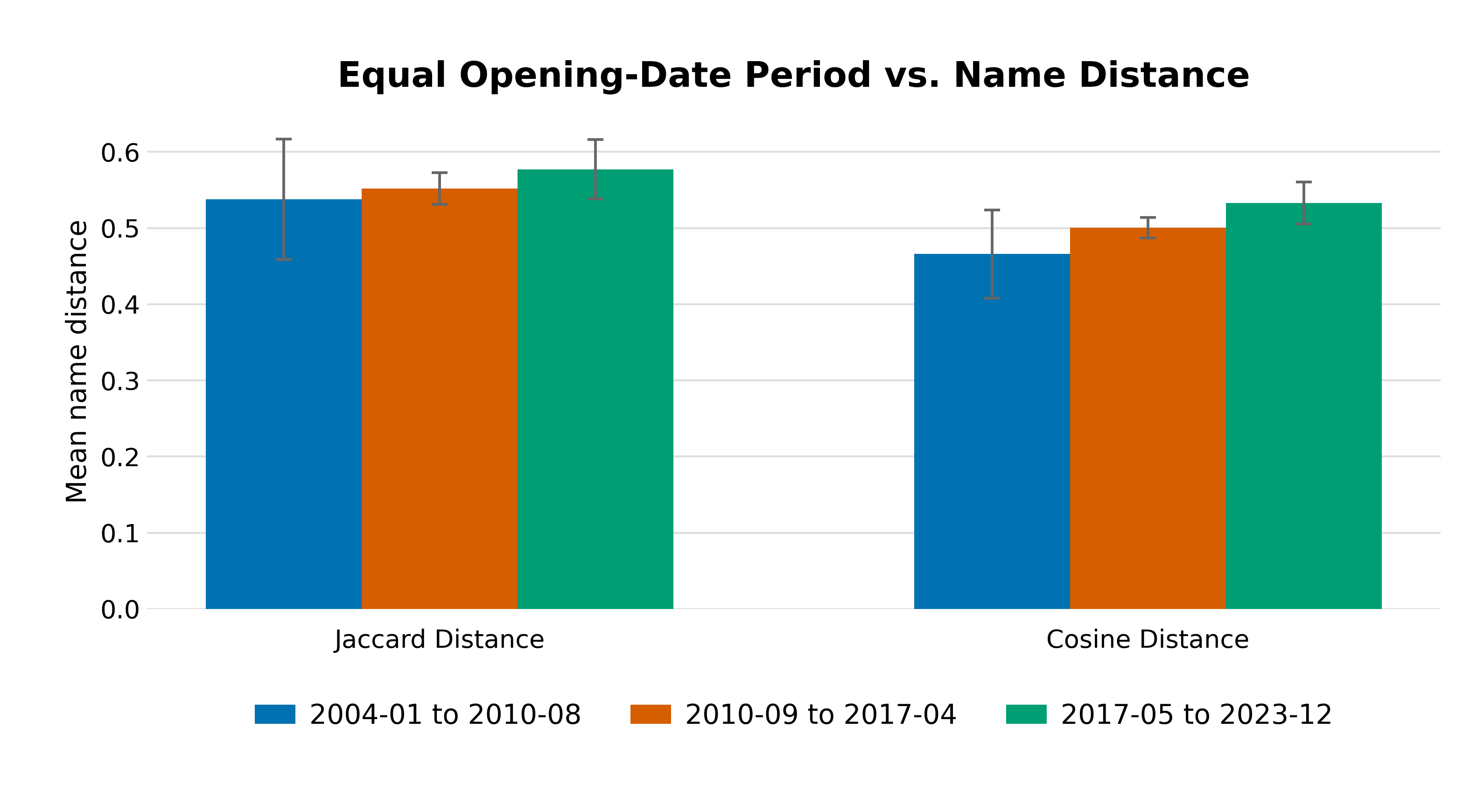}
        \caption{Opening Date}
        \label{fig:opendate_distance}
    \end{subfigure}
    \caption{These three figures show the semantic distance scores between parallel English and Chinese names across different location, Chinese percentage, and opening date groups. Jaccard distance is the frame-based $FD$ score and Cosine distance is translation-based embedding cosine distance with \texttt{all-mpnet-base-v2}.}
    \label{fig:factors_distance}
\end{figure*}

Location is an important factor shaping naming practices. We quantify semantic differences between parallel Chinese and English names using both frame-based Jaccard distance and embedding-based cosine distance. As shown in Figure \ref{fig:location_distance}, ANOVA tests show significant location differences for both metrics. Post-hoc Tukey HSD tests indicate that urban ACRs have significantly more divergent parallel names than rural ACRs for both metrics, while the urban--suburban and rural--suburban differences are not statistically significant. We conjecture that this pattern reflects the different audience structures of urban restaurant markets: in urban areas, where Chinese-speaking customers are more concentrated, ACRs may have stronger incentives to preserve distinctive Chinese-facing names while using more familiar English-facing names for broader local audiences.

The interview study provides a concrete example of the patterns shown in Figure~\ref{fig:location_distance}. 
In urban areas such as Boston's Chinatown, where ACRs are abundant, owners face intense competition and strive to distinguish their establishments.
For instance, R7 shared that a restaurant in Chinatown has to be \emph{``unique to stand out so that they can have business. There is a lot of competition for the same type of food, and other restaurants have been open for longer.''}
R7 identified a type of Chinese food that was not being sold in Chinatown and incorporated distinctive elements of his cuisine into both the Mandarin and English names to attract patrons. 
Conversely, in rural and suburban settings with fewer ACRs, restaurant owners may prioritize personal connections and autonomy over differentiation.
R9 decided to open a restaurant in a rural town to \emph{``escape Chinatown.''}
He and his partner picked their current location because it is \emph{``adventurous''}, and they wanted to \emph{``start a new life with more freedom.''}
The newfound freedom is reflected in R9's choice of naming the restaurant in memory of his daughter and deciding not to create a Mandarin name.


\subsection{Chinese Ethnic Percentage}

We also test whether naming practices vary with local Chinese population share. Using 2022 ACS county-level estimates from NHGIS\footnote{\url{https://www.nhgis.org/data}}, we divide counties into high- and low-Chinese-percent groups based on the sample median. As shown in Figure \ref{fig:chinese_distance}, ACRs in high-Chinese-percent counties show greater frame-based name divergence than those in low-Chinese-percent counties (Jaccard distance: 0.594 vs. 0.539; Welch's $p = 0.005$). However, the continuous Pearson correlation with Chinese percentage is not statistically significant, and embedding-based distance shows almost no group difference (0.517 vs. 0.514; $p = 0.800$). These results suggest that Chinese population share is associated more clearly with categorical naming-frame divergence than with overall embedding distance. One interpretation is that ACR owners in areas with larger Chinese communities may be more comfortable using distinctive Chinese-facing naming frames while keeping English-facing names legible to broader local audiences.

The interviews show that owners strategically consider customer groups, especially age and cultural background, as their customers change over time. For example, R10 adapts food to a rural local population that is less than 1\% Chinese, while R5 targets international Chinese students visiting Chinatown.

This customer awareness often shapes naming choices. Mandarin names help attract Chinese-speaking customers: R2, whose customers are about 50\% Chinese, chose a Chinese name that signals regional dishes and noted that the mismatch between the Chinese and English names sparks curiosity among Chinese customers, who \emph{``want to investigate and try it.''} English names, by contrast, are often chosen to be familiar and legible to American customers. R2 chose the English name first because the \emph{``name tells [American customers] it’s an Asian food.''}

\subsection{Opening Date}
We also investigate whether the opening date is associated with the naming practice. Since it is difficult to obtain the opening date of these restaurants, we use the date from the earliest review for each restaurant on Google Maps to approximate the opening date. To make the periods as comparable as possible, we divide the dates into three equal-month periods: the first period ``2004-01 to 2010-08'', the second period ``2010-09 to 2017-04'', and the third period ``2017-05 to 2023-12''. We find that ACRs in the third period have the highest mean divergence under both Jaccard and Cosine distance. However, the updated ANOVA results are significant only for Cosine distance, and the post-hoc Tukey HSD tests do not show significant pairwise differences between periods. 
To avoid potential bias introduced by manually selecting time intervals, we directly calculated the Pearson correlation using ``days since opening'' to examine whether a temporal trend exists. We found a negative correlation between opening time and Cosine distance using Pearson correlation ($ r(1703) = -.062, p = .010$),  suggesting that newer restaurants tend to have slightly more divergent parallel names; the corresponding Jaccard-distance association is not statistically significant. Our hypothesis is that in recent years ACR owners may have become more comfortable marking their Chinese identity using distinctive Chinese names while keeping familiar English names for local customers, though the updated temporal evidence is weaker than before.
Though most interviewees opened restaurants before 2011, several examples suggest how opening period shaped naming decisions. R4 named a 2003 restaurant after a popular late-1990s movie featuring Chinese characters, explaining that \emph{``it is good to have a simple name to be more acceptable by customers.''} R6, whose restaurant opened in 2010, chose an English name associated with a popular Chinese dish that was receiving media attention, which \emph{``made American customers curious.''} In contrast, the two owners who opened restaurants after 2016 emphasized personal identity: R1 wanted the name to \emph{``be personal''} by including her Chinese name, while R7 said he \emph{``does not care if other people like the name''} when creating and copyrighting it.

Owners also reflected on broader changes in ACRs. R3 confirms the pattern shown in Figure~\ref{fig:opendate_distance}, sharing that \emph{``many new restaurants with creative names are opening. Many owners are now international students or joint ventures.''}, though many also \emph{``opened and closed quickly.''} R10 observed that \emph{``more authentic Chinese food is getting here,''} while R8 argued that \emph{``Chinatowns are dying and spreading to different communities,''} linking this diffusion to greater confidence, English fluency, and integration into U.S. society.

\subsection{Effect size of socio-contextual factors}
To evaluate how much socio-contextual factors influence naming discrepancy, we performed multivariate regression followed by non-linear robustness checks (Appendix \ref{appendix:continuous-duration}). The multivariate regression results show that only rural location ($\beta = -.168, p = .014$) for Jaccard distance and opening date ($\beta = -.018, p = .006$) for Cosine distance yielded statistically significant results. For both naming discrepancy measurements, the multivariate regression models showed low explanatory power ($R^2 < 0.008$). Random Forest regressor robustness check yielded non-positive explanatory power, suggesting no meaningful non-linear relationships are left undetected.
We offer the following reflections on these experiment results. First, we acknowledge a pronounced class imbalance in our dataset, where rural entries account for only 2\% of the sample compared to 91\% for urban entries. However, this class imbalance is not caused by our sampling procedure, but by the data incompleteness in the demographic data source. Furthermore, the low explanatory power of our regression models indicates that restaurant naming cannot be adequately explained by a few macro-level factors; rather, owners' choices are driven by more complex, idiosyncratic decision-making processes.

\section{Conclusion}
In conclusion, our paper studies cultural assimilation through discrepancies between parallel English and Chinese names of ACRs. We adapt an eight-dimensional name-frame schema and create the \textsc{NameFrames} dataset on 884 pairs of ACR names. Using GPT-5.5 labels with human overrides, we find that English names more often emphasize \textit{Location} and \textit{Specialty}, while Chinese names more often emphasize \textit{Positivity}, and to a lesser extent \textit{Culture} and \textit{Ambiance}. Applying the labels to the 1,722 ACRs with both English and Chinese names, we show that these patterns extend beyond the annotated sample. These results suggest that English-facing names foreground legible market information, whereas Chinese-facing names preserve more auspicious and culturally meaningful content.

We further show that name divergence is associated with context, though the strength of the relationship depends on the distance metric. ACRs in urban areas show greater Chinese--English name divergence than rural restaurants, while suburban restaurants generally fall between the two. Counties with higher Chinese population share show larger frame-based differences, but not consistently larger embedding-based differences. Opening-date analyses also suggest that more recently opened restaurants tend to exhibit greater embedding-based divergence, while frame-based results are weaker. Together, these findings suggest that ACR naming practices reflect both market-facing adaptation and the preservation of culturally meaningful Chinese-facing identities, with divergence shaped by local market, ethnic, and temporal context. We complement the computational analysis with ten qualitative interviews with ACR owners to better understand the motivations, audience considerations, and lived experiences behind these naming practices.

\newpage
\section*{Limitations}
 
\paragraph{Names as a single, noisy signal.}
Restaurant names are one public-facing artifact among many, and they are shaped by forces that have little to do with cultural orientation: family history, prior tenants and inherited signage, landlord or municipal sign constraints, trademark availability, delivery-platform display conventions, and simple inertia after ownership changes. A divergent name pair is therefore evidence of a naming \emph{practice}, not a measurement of an owner's identity or degree of assimilation, and we caution against reading our aggregate patterns as individual-level inferences.
 
\paragraph{Recoverability of Chinese names is not missing-at-random.}
Of the 3{,}172 sampled ACRs, we recovered Chinese names for 1{,}722. The remaining 1{,}450 include both restaurants that genuinely have no Chinese name and restaurants whose Chinese signage was simply not visible in available Google Maps photos, Street View imagery, or websites. These two cases are systematically confounded with the very covariates we study: urban restaurants and those in high-Chinese-share counties are more likely to be photographed, to have interior menu photos, and to display Chinese signage prominently. Our analyses condition on successful recovery, so location and ethnic-share effects may be partly attributable to differential visibility rather than differential naming.
 
\paragraph{Orthographic and dialectal flattening.}
We instructed annotators to transcribe into simplified characters. Many ACRs, particularly those founded by Cantonese- and Toisanese-speaking immigrants, use traditional characters, and some romanize according to Cantonese rather than Mandarin. Normalizing to simplified Mandarin erases orthographic and dialectal information that is itself a marker of migration cohort and regional origin, and our \textit{Romanization} frame is defined in ways that implicitly privilege Mandarin pinyin.
 
\paragraph{Frame schema and annotation.}
Our eight frames are adapted from a schema developed qualitatively on 423 English names in a single county, then generalized to Chinese. The resulting categories may not carve Chinese naming practice at its joints: conventions such as the
\begin{CJK}{UTF8}{gbsn}记\end{CJK} suffix, register distinctions among \begin{CJK}{UTF8}{gbsn}轩\end{CJK}/\begin{CJK}{UTF8}{gbsn}楼\end{CJK}/\begin{CJK}{UTF8}{gbsn}阁\end{CJK}/\begin{CJK}{UTF8}{gbsn}坊\end{CJK}, and numerological auspiciousness
are absorbed into broad frames rather than modeled directly. Labels are binary and unweighted, so a name that gestures faintly at \textit{Culture} and one built entirely around a cultural symbol are treated identically. Annotation was performed by co-authors who were not blind to the study hypotheses, which raises the possibility of hypothesis-consistent labeling despite high inter-annotator agreement.
 
\paragraph{Properties of the distance metrics.}
Jaccard distance over eight binary frames is coarse and takes only a small number of discrete values, and the label spaces are asymmetric: \textit{Romanization} is undefined for Chinese names, which mechanically inflates JD for any English name carrying that frame. Cosine distance is computed after a translation step, so translation error propagates directly into the measure; \texttt{all-mpnet-base-v2} is an English-only encoder trained on general text and is not well calibrated for very short proper-noun strings. The two metrics disagree in several analyses (notably Chinese population share), and we report both rather than adjudicating between them, but this also means our conclusions are metric-dependent.
 
\paragraph{Correlational design and inferential caveats.}
All socio-contextual analyses are observational. County-level ACS estimates are aggregates, so associations with Chinese population share are subject to the ecological fallacy: we do not observe the composition of any restaurant's actual customer base. Median-splitting a continuous covariate discards information and can manufacture group differences that the continuous test does not support, which is precisely the pattern we observe. We test three socio-contextual factors under two metrics without correcting for multiple comparisons, so borderline results should be treated as exploratory.
 
\paragraph{Opening date proxy and survivorship.}
We approximate opening date by the earliest Google Maps review, which systematically compresses the histories of older restaurants (Google Maps reviews are sparse before the early 2010s) and misdates restaurants that rebranded, changed ownership, or relocated. Our sample also contains only restaurants operating at collection time, so any temporal trend confounds cohort effects with differential survival---and one of our own interviewees observed that many newer, creatively named restaurants ``opened and closed quickly.''
 
\paragraph{Interview sample and analysis.}
Our qualitative component comprises ten owners in Massachusetts, eight of them urban, six from Fuzhou. Recruitment by walk-in and cold call selects for owners with time and willingness to talk. Because most participants declined recording, our data are English-language notes taken in real time during Mandarin-language interviews, so we have no verbatim transcripts; quoted material reflects a note-taker's contemporaneous rendering rather than the participant's exact words. Themes were identified by authors already familiar with the computational findings.
 
\paragraph{Reliance on proprietary LLMs.}
Frame detection and translation depend on closed, versioned commercial models accessed through APIs. These endpoints are not stable over time, which limits exact reproducibility even with fixed seeds, and we cannot rule out that widely circulated ACR names or prior work on this topic appeared in pretraining data. LLMs also encode culturally situated priors about what counts as ``cultural,'' ``creative,'' or market-legible, and those priors are plausibly Western-centric in ways that align with---rather than independently test---our hypotheses. We therefore treat model outputs as structured approximations validated against human labels on a limited sample, not as ground truth.
 
\paragraph{Scope.}
We study one diaspora, in one country, at one moment. ``Chinese'' is used throughout as an administrative convenience that flattens Cantonese, Fujianese, Taiwanese, Sichuanese, Hui, and Southeast Asian Chinese communities with quite different migration histories and naming conventions. We make no claims about other immigrant cuisines or about ACRs outside the United States.

\paragraph{Missing data in data source}
Our demographic data were sourced from IPUMS NHGIS. However, due to missing data in certain rural areas, we had to drop incomplete records during multivariate regression for effect size analysis. While this ensured complete feature coverage, it introduced class imbalance that potentially impacted data analysis.

\section*{Ethical Considerations}
 
\paragraph{Human subjects and consent.}
Both the transcription task and the owner interviews were reviewed by our institution's IRB and received exempt determination). Interviews were conducted in Mandarin; participants received a verbal description of the study's purpose and data handling and gave verbal informed consent before beginning. Most participants declined audio recording, and we honored this by taking written notes only. Participation was voluntary and participants could decline any question or withdraw at any point. 
 
\paragraph{Crowdworker compensation and conditions.}
Transcription was performed by Prolific workers who self-identified as first-language Chinese speakers. Across the task, workers logged approximately 67 hours. We paid an effective rate of 15 dollars/hour, above the state minimum wage and above Prolific's recommended floor. Surveys were piloted to calibrate time estimates, and we did not reject submissions on the basis of disagreement between the two independent annotators.
 
\paragraph{Privacy and identifiability.}
Restaurants are public commercial entities and their names and locations are publicly listed, so we treat name and location data as public record. Owners, however, are individuals, and small-business owners are readily re-identifiable from a combination of city, opening year, and cuisine. We refer to interviewees only as R1--R10, do not report restaurant names, exact locations, or the names of family members, and paraphrase or omit details from the interviews that would narrow identification to a single establishment---including several personal and family circumstances that participants shared. Some quoted material remains distinctive despite these steps; we retained it only where participants indicated they were comfortable with the detail being reported.
 
\paragraph{Data release.}
We release the \textsc{NameFrames} annotations---parallel name pairs with frame labels---together with our codebook, prompts, and analysis code, to support replication. We do not redistribute raw scraped Google Maps records, review text, photographs, or owner-associated metadata, both to respect platform terms of service and because aggregated business records are more easily repurposed for commercial targeting than the derived annotations we need for replication. No interview notes are released.
 
\paragraph{Framing and risk of essentialization.}
``Cultural assimilation'' is a contested construct with a history of being used normatively to rank groups by their proximity to a dominant culture. Our name-distance measures are descriptive: a larger distance indicates that two names encode different meanings, not that an owner is less integrated, less authentic, or more foreign. We have tried to write against a reading in which divergence scores become an assimilation index applied to ethnic communities. Relatedly, frames such as \textit{Culture} and \textit{Romanization} risk reproducing an exoticizing gaze if read as markers of foreignness rather than as descriptive categories of naming practice.
 
\paragraph{Potential for misuse.}
Aggregated, geolocated data about ethnic businesses can be repurposed for targeting---commercially, in the form of marketing or acquisition pressure, and in worse cases as input to harassment or discrimination. Anti-Asian harassment of restaurant businesses in the U.S. has been documented and recent. This informed our decision to release annotations rather than a geolocated business registry, and we encourage anyone extending this work to weigh the same tradeoff.
 
\paragraph{Positionality.}
The annotation team consists of native Chinese speakers with professional English proficiency, several of whom are themselves immigrants to the United States; the interviews were conducted by Mandarin-speaking members of the team. This shared background aided rapport and interpretation of culturally specific names, and it also means our intuitions about what is ``auspicious,'' ``creative,'' or ``market-legible'' are embedded in the codebook rather than external to it.
\section*{Acknowledgements}
We used AI coding assistants for coding in the statistical experiments in Section 5.5. All AI-generated code underwent two rounds of review by the authors, with manual verification where necessary. The text part in corresponding section was written by the authors. For the empirical/statistical results description, we consulted AI agent exclusively for wording accuracy. The authors retain full responsibility for all code and content in this paper. 

\bibliography{custom, anthology-1, anthology-2}

\newpage
\appendix
\onecolumn

\section{Data Collection}
\label{appendix:data_collection}

\subsection{ACR Statistics}
\label{appendix:acr_stats}
Include the number of ACRs per state with a huge table, including the total number and the sampled number. Also include a randomly picked ACR example along with the meta information we have. 

\begin{longtable}{|c|c|c|}
\hline
\textbf{state}       & \multicolumn{1}{l|}{\textbf{sample\_frequency}} & \multicolumn{1}{l|}{\textbf{national\_frequency}} \\ \hline
California           & 357                                             & 4265                                              \\ \hline
New York             & 260                                             & 2979                                              \\ \hline
Texas                & 165                                             & 1990                                              \\ \hline
Pennsylvania         & 153                                             & 1830                                              \\ \hline
New Jersey           & 130                                             & 1546                                              \\ \hline
Florida              & 123                                             & 1596                                              \\ \hline
Illinois             & 98                                              & 1152                                              \\ \hline
Massachusetts        & 78                                              & 897                                               \\ \hline
Ohio                 & 73                                              & 971                                               \\ \hline
Virginia             & 72                                              & 907                                               \\ \hline
Michigan             & 66                                              & 783                                               \\ \hline
Georgia              & 65                                              & 899                                               \\ \hline
North Carolina       & 65                                              & 931                                               \\ \hline
Washington           & 63                                              & 707                                               \\ \hline
Maryland             & 58                                              & 723                                               \\ \hline
Alaska               & 46                                              & 68                                                \\ \hline
Connecticut          & 45                                              & 526                                               \\ \hline
Hawaii               & 45                                              & 169                                               \\ \hline
District Of Columbia & 43                                              & 124                                               \\ \hline
New Mexico           & 43                                              & 133                                               \\ \hline
Missouri             & 42                                              & 619                                               \\ \hline
Colorado             & 41                                              & 480                                               \\ \hline
Rhode Island         & 41                                              & 150                                               \\ \hline
Arizona              & 40                                              & 464                                               \\ \hline
Idaho                & 40                                              & 117                                               \\ \hline
Louisiana            & 40                                              & 353                                               \\ \hline
Montana              & 40                                              & 75                                                \\ \hline
New Hampshire        & 40                                              & 177                                               \\ \hline
Oregon               & 40                                              & 407                                               \\ \hline
Arkansas             & 39                                              & 199                                               \\ \hline
Indiana              & 39                                              & 474                                               \\ \hline
Maine                & 39                                              & 148                                               \\ \hline
Delaware             & 36                                              & 138                                               \\ \hline
Kentucky             & 36                                              & 294                                               \\ \hline
Nevada               & 36                                              & 287                                               \\ \hline
West Virginia        & 36                                              & 140                                               \\ \hline
Wyoming              & 36                                              & 53                                                \\ \hline
Kansas               & 35                                              & 263                                               \\ \hline
South Dakota         & 35                                              & 55                                                \\ \hline
Utah                 & 35                                              & 200                                               \\ \hline
Vermont              & 35                                              & 82                                                \\ \hline
Alabama              & 34                                              & 370                                               \\ \hline
Minnesota            & 34                                              & 415                                               \\ \hline
Nebraska             & 34                                              & 178                                               \\ \hline
Tennessee            & 34                                              & 453                                               \\ \hline
North Dakota         & 33                                              & 41                                                \\ \hline
Oklahoma             & 33                                              & 306                                               \\ \hline
Iowa                 & 31                                              & 294                                               \\ \hline
Mississippi          & 30                                              & 202                                               \\ \hline
South Carolina       & 30                                              & 489                                               \\ \hline
Wisconsin            & 30                                              & 429                                               \\ \hline
\end{longtable}

\subsubsection{Data Filtering}
\label{appendix:acr_filtering}
We first filtered out any restaurant that has the phrase 'fast food','chain' or 'express' in its google maps subtype, description, name or review. Then we curated a list of well-known chain Chinese restaurants in the U.S. to filter out the remaining chain restaurants that didn't mark themselves as chain.

\subsection{\texttt{NameFrames} Dataset}
\label{appendix:namefr_data}

\subsubsection{ACR Distribution across States}

We categorized each restaurant as urban (2), suburban (1), or rural (0) based on its ZIP code. We developed this categorization using \citet{paykin2021area}'s guidelines for area classification.

\begin{figure}[H]
    \centering
    \includegraphics[width=0.5\linewidth]{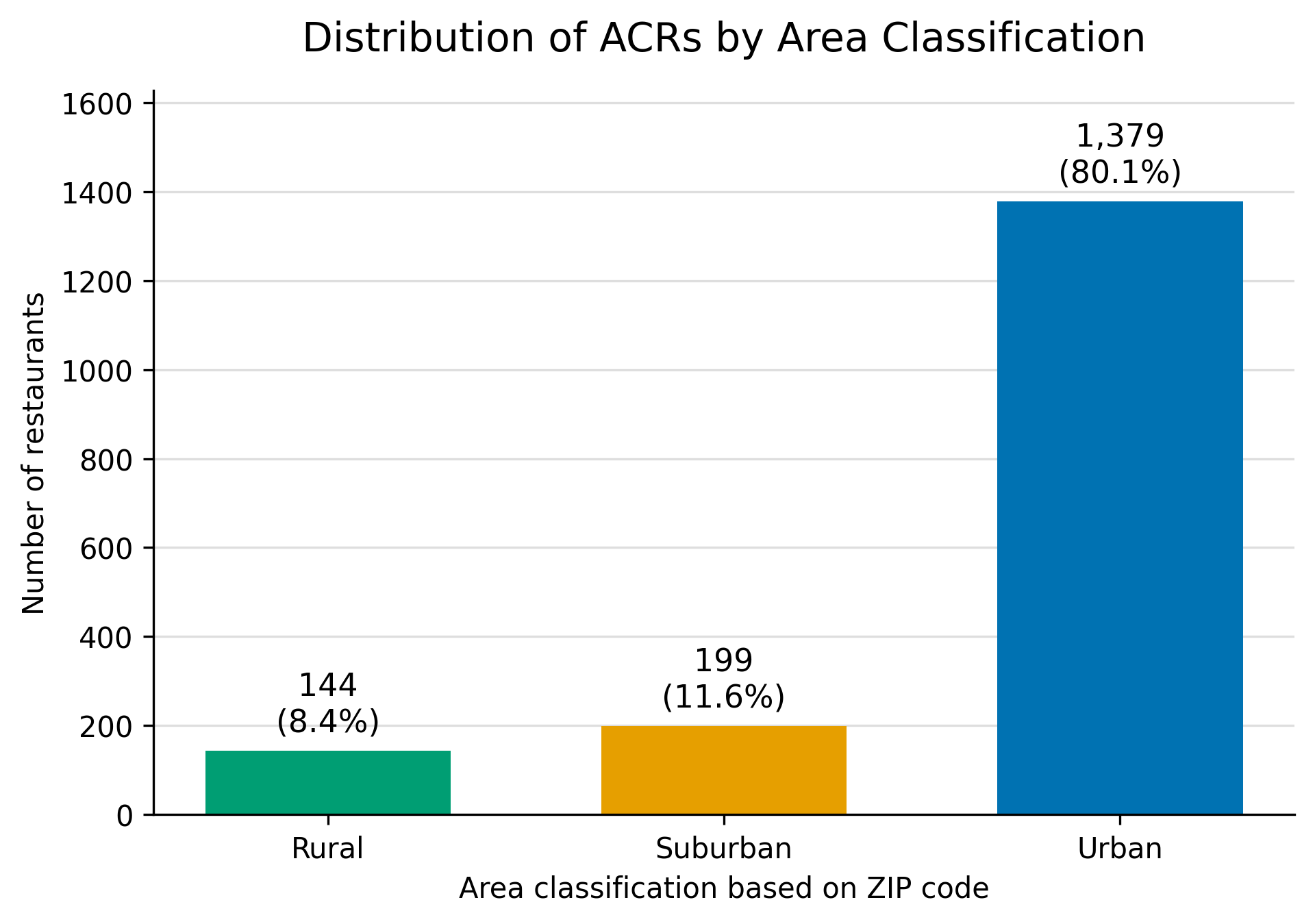}
    \caption{Caption}
    \label{fig:acr-area-distribution}
\end{figure}

\subsection{Codebook}
\label{appendix:codebook}

Here are the specific changes we have made to \citet{chen2018authenticity}'s 9 name frames including (1) \textit{Personal Names}, (2) \textit{Speciality}, (3) \textit{Superlatives and Other Positive Modifiers}, (4) \textit{Cultural Association}, (5) \textit{Place Names}, (6) \textit{Ambiance}, (7) \textit{Others}, (8) \textit{Transliterated Hanzi}, (9) \textit{Express Takeout}. First, we renamed (3) to \textit{Positivity}, (4) to \textit{Culture}, (5) to \textit{Location}, and (8) to \textit{Romanization} for simplicity in language. Second, we added a rule to include intimacy words related to family or relationship to the \textit{Ambiance} frame, since we found that many ACRs like to use these words during our annotation. Third, we replaced the \textit{Others} frame with a more specific \textit{Creativity} frame and extended its definition to names with creative puns such as ``Wok N' Roll''. Fourth, we removed (9) in our analysis because the names of express ACRs tend to be highly repetitive and we excluded them in our analysis.

\subsubsection{English}
    \begin{itemize}
      \item \textbf{Personal names}: usually, but not necessarily, surnames or first names. 
Here are some common examples:\begin{itemize}
    \item Surnames with a possessive: \textit{Qing’s Kitchen}, \textit{Hoy’s Wok}, \textit{P.F. Chang’s China Bistro}
    \item Names without a possessive: \textit{China Lee}, \textit{Hunan Mao}, \textit{House of Louie}, \textit{Mei Mei Dumplings}
    \item Surnames with a form of address or courtesy title: \textit{Mr Chow}, \textit{Madame Wu’s Asian Bistro}, \textit{MaMa Lu’s Dumpling House}
    \item Nicknames for family members or acquaintance: \textit{Ah-X}, \textit{Liu Po}
    \item English names: \textit{Paul’s Kitchen}, \textit{House of Benny}, \textit{Bebe Fusion}, \textit{Sam’s Chinese BBQ}
\end{itemize}
      \item \textbf{Specialty}: refers to certain food or dishes for which some restaurants are known for. Here are some common examples:
      \begin{itemize}
        \item Names with special meat, seafood, and tofu: \textit{Duck Queen, Fishball House, Little Fat Sheep, Seafood Village, Tofu House}
        \item Names with BBQ and buffet: \textit{Sam’s Chinese BBQ, New Panda Buffet}
        \item Names of tea-drinking restaurants: \textit{Green Tea House, Boba Tea Station, Tea Time}
        \item Names with Dim Sum, Yum Cha, and Dim Sum dishes: \textit{Dim Sum Express, Yum Cha, Dumpling House, Tasty Bao, Egg Roll King}
        \item Names with hot pot and pot: \textit{Mongolian Hot Pot, China Pot King}
        \item Names with rice, noodles, congee: \textit{Noodle House, Rice N Noodles, The Congee, Porridge House}
        \item Names with religious meanings, vegetarian, and other cultures: \textit{China Halal, Happy Veggie Garden, Buddha’s Feast}
        \item Names with wok, snacks: \textit{China Wok, Rice Wok, Beijing Snacks}
    \end{itemize}
      \item \textbf{Superlatives and Other Positive Modifiers}: refers to using superlatives and positive words in the names. Here are some examples:\begin{itemize}
            \item Names with a superlative: \textit{China Best Express, China Great Buffet, Perfect Chinese Food, Elite, The Legendary, No. 1 Noodle House}
            \item Names with modifiers denoting superb taste: \textit{Delicious Food Corner, Tasty Wok Cuisine, Yummy Cafe, Luscious Dumplings, Chinese Gourmet Kitchen, Fresh China Kitchen}
            \item Names with gold, golden, silver, diamond, jade: \textit{Gold Medal Chinese Food, Golden China, Silver Palace, Shanghai Diamond Garden}
            \item Names with modifiers denoting good luck such as 8: \textit{888 Fast Food, Lucky Seafood}
            \item Names with words that elicit positive expectations: \textit{Happy Family, House of Joy, Sunny Palace, Pleasure Ocean, Nice Time Cafe}
            \item Names with words that exude opulence: \textit{Grand Buffet, Feast from the East, Jumbo Chinese Fast Food}
            \item Names with royalty implications: \textit{China King, King Fu, Kingchops, Empress Pavilion, Empire Palace, Regent China Inn, Royal Spring, VIP, The Dynasty}
            \item Names with OK: \textit{OK Chinese Food}
            \item Names with red: \textit{Red House}
        \end{itemize}

      \item \textbf{Ambiance}: refers to a kind of atmosphere which makes customers feel comfortable and enjoy the dining experience, or it often indicates family or friendship, or being homesick, which shortens the psychological distance between the restaurant and the patrons. Here are some common examples:
        \begin{itemize} 
            \item Names with the word “garden” that conveys a sense of tranquility and leisurely refinement: \textit{Garden and Ding’s, Mandarin Garden, Tasty Garden, New Garden}
            \item Name with the word “palace,” which suggests the image of imperial-style fine dining: \textit{China Palace, Rowland Palace, Silver Palace}
            \item Names that describe home-cooked meals in a more casual setting with words like “house,” “home,” “kitchen,” or “dining room”: \textit{House of Benny, Joy House, Moon House, New RJ Chinese Kitchen, Tao Chinese Kitchen, Chinese Kitchen, Spicy Home, Yang’s Dining Room}
            \item Names that evoke images of a simpler life or country living, like “hut”, “village”, “inn”, “bistro”: \textit{Boba Tea Hut,  Green Village, Hunan Village,  Bamboo Inn, Plum Tree Inn, Wok Inn,  Madame Wu’s Asian Bistro, New Canton Bistro}
            \item Names that contain family members: \textit{Brother's Chinese Restaurant, Uncle Tom's Hot Pot, Grandma’s Pancake, Mama’s Kitchen}
            \item Names suggesting friendship: \textit{Friend’s house, Friends}
            \item Names that remind patrons of their hometown: \textit{Hometown Szechuan, Home Chinese Cafe}
        \end{itemize}
      \item \textbf{Creativity}: refers to names that contain puns or words used creatively. Here are some examples:
        \begin{itemize}
            \item "wok with me" (sounds like “walk with me”), "pho king good" (sounds like “fucking good”), "rice to meet you" (sounds like “nice to meet you”), "Wok N Talk" ('wok' ryhmes with 'talk'), etc
        \end{itemize}
      \item \textbf{Culture}: refers to the Chinese, American, or other cultural symbols and references that appear in the names of restaurants. Here are some common examples:

        \begin{itemize}
            \item Names that contain animals or mystical creatures native to China: \textit{Panda House, Dragon Garden, Phoenix tavern}
            \item Names that contain famous historical monuments: \textit{Great Wall, Forbidden City Restaurant, Golden Gate}
            \item Names that contain plants native to China: \textit{Bamboo Place, Plum Blossom}
            \item Names that contain famous historical or well-known figures: \textit{Jackie Chen’s Place}
            \item Names that contain the name of a specific Dynasty name: \textit{the Tang Dynasty, the Song Dynasty}
            \item Names that contain cultural artifacts: \textit{Lantern Garden, Silk House}

        \end{itemize}
      \item \textbf{Location}: Names that refer to Asia, China, or specific regions in China or other places. Here are some common examples:
        \begin{itemize}
            \item Names that refer to China, Asia, or other Asian countries: \textit{Chinese restaurant, Mandarin Garden, Asia Buffet, Chinese and Thai cuisine, Cathay House}
            \item Names that refer to a specific province or city: \textit{Xi’an Noodles, ShanDong Garden, Beijing House, Hong Kong restaurant, Shanghai and Tokyo, Canton Bistro}
            \item Names that contain the name of a town or street where the restaurant is located in: \textit{Wellesley Square sushi buffet, Brookline Wok}

        \end{itemize}
      \item \textbf{Romanized}: containing Romanized Chinese characters that are not names of a person or names that indicate relationships. Here are some common examples:
        \begin{itemize}
            \item Examples: \textit{Lu Fu Zi, Wei Shi Jia, Shu Da Xia restaurant, Hai Di Lao Garden, Din Tai Fung, Yi Mei Deli}
        \end{itemize}
    \end{itemize}
\subsubsection{Chinese}
\textbf{Original Chinese}

\begin{CJK*}{UTF8}{gbsn}
    \begin{itemize}
        \item \textbf{人名}: 包括姓氏，名字，全名或者昵称。下面是一些例子:
            \begin{itemize}
                \item 餐馆名带有姓氏：林记烧腊，李记面馆，王家园，张师傅厨房
                \item 餐馆名带有名字：亚轩十八，威威，嘉嘉海鲜菜馆
                \item 餐馆名带有全名：李明餐厅，刘一手火锅
                \item 餐馆名带有昵称：六婆串串香，阿庆东北饺子馆
            \end{itemize}
        \item \textbf{特色}: 一般指带有一个餐馆名有特点的菜系，食物类型或者饮食形式。下面是一些例子：
            \begin{itemize}
                \item 餐馆名带有肉类，海鲜，豆腐种类：北京烤鸭餐厅，小肥羊，海鲜渔村，韩式豆腐汤
                \item 餐馆名带有烧烤或者自助：老村长烧烤屋，巴西烤肉，老友撸串店，金钱豹自助餐厅
                \item 餐馆名带有饮茶奶茶：时光奶茶
                \item 餐馆名带有点心：翠华茶餐厅，乐茶轩，一点心
                \item 餐厅名带有包子饺子：南翔小笼包，东北饺子馆
                \item 餐厅名带有火锅或者特色锅煲：蒙古火锅，砂锅居，成都冷锅，韩国海鲜豆腐煲
                \item 餐厅名带有米饭和面粉：小明炒饭，湖南米粉，绿色吴家粥铺，嘉和粥点，碰记热干面
                \item 餐厅名带有伊斯兰特色，素食标签，或者其他文化标签：清真拉面，素食天地，阿曼尼萨汗美食城
                \item 餐厅名带有小吃：天津特色小吃，江浙小吃

            \end{itemize}
        \item 最高级和其他积极的修饰语: 指代使用比较级，最高级的修饰语还有使用积极正能量的修饰词。以下是一些例子：
            \begin{itemize}
                \item 餐馆名指代好以及其比较级最高级的单词：京城第一，完美中餐厅，最好拉面店，顶好餐厅
                \item 餐馆名有指代味道好的修饰词：美味家园，好吃小馆，鲜芋仙，香喷喷厨房
                \item 餐馆名有金，银，玉，翡，翠，鼎，御, 宝等这些: 金山餐厅，聚宝盆，玉园，东方明珠
                \item 餐馆名有好运的修饰词比如发，8，八，好运：好运楼，发祥地，88酒家
                \item 餐馆名可以激起美好的期待：快乐家庭，阳光宫殿，好时光咖啡，圆满小厨，向日葵餐厅
                \item 餐馆名指代丰富富裕比如富，八，发，祥，嘉，聚，旺，和：盛大自助餐，富国餐厅，东方盛宴
                \item 餐馆名指代皇家和尊贵：御尚坊，帝国餐厅，皇家特供，王朝餐厅，王品牛排
                \item 餐馆名包含OK：OK拉面
                餐馆名包含专业程度：大厨厨房，匠人厨房
                \item 餐馆名包含积极的意向比如红色，升日，花好月圆：红房子，日升餐厅

            \end{itemize}
        \item 氛围: 包括暗示环境氛围的词语，一般作为地点的高级表达方式，给顾客一种舒适的体验。以下是一些例子：
            \begin{itemize}
                \item 餐馆名包含餐厅和厨房的高级表达方式：：南北小馆、新和小馆、食苑小厨、私房小厨、司膳房、御膳房
                \item 餐馆名包含古代高级地点词汇：美食轩、金鼎轩、凌波轩、湘德楼、海鲜酒家、川坝子酒楼、峨眉酒家、听涛阁、逐帆亭、望海楼、拾贝坊、采菱榭、御风堂、浣沙居
                \item 餐馆名包含带有艺术表达气息的地点名称：楼上花园、御香园、万秀园餐厅、熙春园餐厅

            \end{itemize}
        \item 双关等创意题目: 包括双关、谐音等等各种有创意的修辞方法。以下是一些例子：
            \begin{itemize}
                \item 餐馆名包含双关或者谐音：四面八方（含有“面”，适合当面馆的名字），海食蜃楼（取自成语“海市蜃楼”），与食俱进（取自成语“与时俱进”），美食美客（取自成语“每时每刻”），天外飞鲜（取自成语“天外飞仙”）
                \item 餐馆名包含文化：老友记（电视剧名，也泛指老友相聚）、三味真火（“三味”可以指世间一切美味，也是红孩儿的看家本领；“真”字有真诚、真实的意思；“火”字寓意生意红火）、楚留香（本是人名，同时“留香”也意味着留下香气）

            \end{itemize}
        \item 文化: 指餐厅名字中含有中华文化或美国文化符号或代表。以下是一些例子：
            \begin{itemize}
                \item 餐馆名包含动物：熊猫园，龙酒家, 凤凰茶楼
                \item 餐馆名包含历史古迹：长城饭店，天坛，金门
                \item 餐馆名包含植物：松鹤楼，梅花楼，绿竹园
                \item 餐馆名包含著名历史人物：小平饭店，秦始皇大酒楼
                \item 餐馆名包含朝代名称：唐朝饭店
                \item 餐馆名包含工艺品: 大红灯笼，青瓷园
            \end{itemize}
        \item 地点: 名字中含有与中国，亚洲，或其他地区、城市、街道的名字。以下是一些例子：
            \begin{itemize}
                \item 餐馆名包含中国或亚洲：中国餐厅，亚洲楼，华园，华明，中
                \item 餐馆名包含城市名：北京园，奉天小馆，川湘味，金粤园
                \item 餐馆名包含其他地方或街道名：曼哈顿小吃, 芬威楼

            \end{itemize}
    \end{itemize}
\end{CJK*}
\textbf{English Translation}
    \begin{itemize}
        \item Personal name: includes surname, first name, full name, or nicknames. Here are some examples:
            \begin{itemize}
                \item restaurant names containing surnames: \textit{Lin's shaola, Lee's Noodle restaurant, Wang's garden, Chef Zhang's Kitchen}
                \item restaurant names come with first names: \textit{Asia Pavillion 18, Weiwei, Jiajia Seafood Restaurant}
                \item restaurant names with full name: \textit{Li Ming Restaurant, Liu Yishou Hotpot}
                \item restaurant names with a nickname: \textit{Liu Po Delicious Skewers, Ah Qing Northeast Dumpling House}
            \end{itemize}
        \item Specialty: Generally refers to the cuisine, food types or dietary forms that are unique to a region or restaurant. Here are some examples
            \begin{itemize}
                \item Restaurant names include types of meat, seafood, and tofu:
                \textit{Beijing Roast Duck Restaurant, Little Sheep, Seafood Fishing Village, Korean Tofu Soup}
                \item Restaurants with barbecue or buffet in their names: \textit{Old Village Chief’s BBQ House, Brazilian BBQ, Old Friend’s Skewers, Golden Leopard Buffet Restaurant} 
                \item restaurant names containing Yumcha or milk tea: \textit{Time Milk Tea}
                \item Restaurant names indicating dim sum: \textit{Tsui Wah Tea Restaurant, Happy Tea House, One Dim Sum}
                \item Restaurant names with steamed buns and dumplings: \textit{Nanxiang Steamed Dumplings, Northeast Dumpling House} 
                \item restaurant names containing hot pot or special pot: \textit{Mongolian hot pot, clay pot, Chengdu cold pot, Korean seafood tofu pot} 
                \item Restaurant names with rice and flour: \textit{Xiaoming Fried Rice, Hunan Rice Noodles, Green Wu's Porridge Shop, Jiahe Porridge and Snacks, Pengji Hot Dry 
                Noodles} 
                \item restaurant names with religous indications, vegetarian labels, or other cultural labels: \textit{Halal Ramen, Vegetarian World, Oman 
                Nisa Khan Food City}.
                \item restaurant names containing snacks: \textit{Tianjin special snacks, Jiangsu and Zhejiang snacks}
            \end{itemize}
        \item Superlatives and other positive modifiers: refers to the use of comparatives, superlative modifiers and modifiers that use positive energy. Here are some examples:
            \begin{itemize}
                \item Restaurant names refer to good and its comparative superlative words: \textit{No. 1 in Beijing, Perfect Chinese restaurant, Best Ramen Shop, Top Good Restaurant} 
                \item restaurant names with modifiers that refer to good taste: \textit{Delicious Home, Delicious Bistro, Fresh Taro Fairy, Fragrant Kitchen}
                \item  Restaurant names include gold, silver, jade, jade, emerald, tripod, imperial, treasure, etc.: \textit{Jinshan Restaurant, Cornucopia, Jade Garden, Dongfang Ming
                beads} 
                \item restaurant names with good luck modifiers such as 8, Good Luck: \textit{Good Luck Building, Lucky Origin, 88 Restaurant}
                \item  Restaurant names that can arouse good expectations: \textit{Happy Family, Sunshine Palace, Good Time Coffee, Perfect Kitchen, Sunflower
                Restaurant}
                \item  Restaurant names refer to richness and opulence such as "rich", "fortune", "lucky", "beautiful", "Union", "prosperous", and: Shanda Buffet, \textit{Rich Country Cuisine 
                Hall, Oriental Feast} 
                \item  Restaurant names refer to royalty and nobility: \textit{Royall Kitchen, Empire Restaurant, Royal Special, Dynasty Restaurant, King Steak} 
                \item restaurant names containing "OK": \textit{OK Ramen} \item restaurant name contains professional level: \textit{Chef Kitchen, Artisan Kitchen}
                \item  Restaurant names containing positive connotations such as red, rising sun, flowers and full moon: \textit{Red House, Sunrise Restaurant}
            \end{itemize}
        \item Atmosphere: Including words that imply the atmosphere of the environment, generally used as a high-level expression of a place to give customers a comfortable experience. Here are some examples:
            \begin{itemize}
                \item Restaurant names include advanced expressions of restaurants and kitchens: \textit{North and South Bistro, New Peace Bistro, Snack Garden Kitchen, Private Kitchen, ROyal Kitchen, Imperial Kitchen}
                \item The names of restaurants include ancient high-end location words: \textit{Delicacy Pavillion, Golden Ding Pavillion,WavesHouse, Hunan Virtue Building, Seafood Restaurant, Chuanbazi Restaurant, Emei Restaurant, Hearing Waves Pavilion, Chasing Boat Pavilion, Seaview Tower, Shell Picking Bistro, Lotus Picking Pavilion, Royal Wind Hall, Gauze Washing Home}
                \item Restaurant names include place names with artistic expression: \textit{Upstairs Garden, Royal Fragrance Garden, Million Beauty Garden Restaurant, Bright Spring Garden Restaurant}
            \end{itemize}
        \item Creative topics such as puns: including puns, homonyms and other creative rhetorical methods. Here are some examples:
            \begin{itemize}
                \item  Restaurant names contain puns or homophones: \textit{Four Directions and Eight Places} (In the Chinese idiom Si Mian Ba Fang, whose literal translation if "Four Directions and Eight Places", the character of Mian is the same as the character for noodle, hence making the name suitable for a noodle restaurant), \textit{Food Mirage} (taken from the Chinese idiom "mirage"), \textit{Advance with Food} (taken from the Chinese idiom "advance with the times"), \textit{Gourmet food and delicious food} (Romanization "Mei Shi Mei Ke") (from the idiom "every moment"(Romanization "Mei Shi Mei Ke"). The Chinese characters in the idiom are replaced with characters of similar pronunciation in the restaurant name), \textit{Freshness from the Sky} (from the Chinese idiom "Tianwai Feixian")
                \item restaurant names containing cultural or media symbols: \textit{Friends} (the name of the TV series, also generally refers to the gathering of old friends), \textit{Three Zhen Fires} ("Sanwei" can refer to all the delicious food in the world, and is also Honghaier's special skill; the word "zhen" means sincerity and authenticity ;The word "fire" means business is booming), \textit{Chu Liuxiang} (originally a person's name, and "Liuxiang" also means leaving a fragrance)
            \end{itemize}
        \item Culture: Refers to restaurants whose names contain symbols or representatives of Chinese culture or American culture. Here are some examples:
            \begin{itemize}
                \item Restaurant names include animals: \textit{Panda Garden, Dragon Restaurant, Phoenix Tea House}
                \item The restaurant name includes historical monuments: \textit{Great Wall Hotel, Temple of Heaven, Golden Gate}
                \item The restaurant name contains plants: \textit{Songhe Tower, Plum Blossom Tower, Green Bamboo Garden}
                \item The restaurant names include famous historical figures: \textit{Xiaoping Hotel, Qin Shihuang Restaurant}
                \item The restaurant name contains the name of the dynasty: \textit{Tang Dynasty Hotel}
                \item The restaurant name contains handicrafts: \textit{Red Lantern, Celadon Garden}
            \end{itemize}
        \item Location: The name contains names related to China, Asia, or other regions, cities, and streets. Here are some examples:
            \begin{itemize}
                \item restaurant names containing words like "Chinese" or "Asian": Chinese Restaurant, Asia House, China Garden, China Bright
                \item  restaurant names include city names: Beijing Garden, Fengtian Restaurant, Sichuan and Hunan Flavor, Jinyue Garden
                \item Restaurant names contain other place or street names: Manhattan Eats, Fenway House
            \end{itemize}
    \end{itemize}

\setlength{\extrarowheight}{3pt}
\begin{table}[ht!]
    \centering
    \footnotesize
    \begin{tabular}{|c|cccc|}
            \Xhline{2\arrayrulewidth}
             & \multicolumn{2}{c}{\textbf{Chinese}} & \multicolumn{2}{c|}{\textbf{English}} \\
            \Xhline{1\arrayrulewidth}
             & Cohen's $\bar{\kappa}$ & Fleiss $\bar{\kappa}$ & Cohen's $\bar{\kappa}$ & Fleiss $\bar{\kappa}$ \\
            \Xhline{2\arrayrulewidth}
\textit{Personal Name}         & 0.8235                      & 0.8240               & 0.812                      & 0.8119               \\
\textit{Specialty}    & 0.9034                      & 0.9032               &  0.8294                     & 0.8297               \\
\textit{Positivity}   & 0.7989                      & 0.7989               &  0.8011                     & 0.8002               \\
\textit{Culture}      & 0.8387                      & 0.8381               &  0.8545                    & 0.8542              \\
\textit{Location}     & 0.8856                      & 0.8858              & 0.935                      & 0.935               \\
\textit{Ambiance}     & 0.8788                      & 0.8787               & 0.8331                    & 0.8337              \\
\textit{Creativity}   & 0.8935                      & 0.8928              & 0.8543                      & 0.8541               \\
\textit{Romanization} & -                           & -                    & 0.7822                      & 0.7841     \\         
            \Xhline{1\arrayrulewidth}
\end{tabular}
\caption{This table shows the inter-annotator agreement (IAA) scores in Cohen's Kappa and Fleiss Kappa on the English and Chinese names of 1000 ACRs. Romanization is not applicable to Chinese names, so we used ``-''. }
\label{tab:annotation_agreement}
\end{table}

\subsubsection{Frame Annotation Agreement}

Table \ref{tab:annotation_agreement} reports the inter-annotator agreement (IAA) scores among three annotators for each frame in Cohen's Kappa and Fleiss Kappa on the English and Chinese names of 884 ACRs. Overall, we observe high IAA scores across 8 different frames.

\section{Models}
\label{appendix:models}

\subsection{Rule-based Method}
\label{appendix:rule_model}

For most of the categories, we developed a list of vocabulary for each and labeled the restaurant names with regex. Here are the lexicons.\\
\begin{itemize}
    \item Chinese
    \begin{CJK*}{UTF8}{gbsn}
        \begin{itemize}
            \item \textbf{Location}:``中'',"亚","华","北京","香港","上海","川","湘","陕","山东","粤","广东","杭州","南京","西安","重庆","成都","桂林"
            \item \textbf{Culture}:"熊猫","龙","凤","长城","天坛","金门","松","鹤","梅花","竹","小平","秦始皇","汉","唐","宋","灯笼","瓷","丁香","荷","菊","功夫","丝绸"
            \item \textbf{Ambiance}:"小馆","厨","膳房","轩","楼","阁","亭","坊","榭","堂","居","园","家","宫","庭","村","乡",\\"屋","苑","吧","城","府","堂","屯","邨","庄","湾"

            \item \textbf{Specialty}:"烤","羊","牛","猪","鸡","鸭","鹅","海鲜","豆腐","串","自助","茶","点心","包子","饺","锅","煲","饭","粉","粥","面","清真","素食","小吃","馍","烧烤","寿司","腊","卤","辣","饼","烫","烩","鱼","米","渔","汤"
            \item \textbf{Positivity}:"第一","明","美","喜","好","鲜","香","金","银","玉","翡","翠","鼎","御","宝","八","发",\\"8","福","乐","满","祥","嘉","聚","旺","和","盛","富","御","帝","皇","王","鸿","冠","太子","怡","佳","兴","升","锦","雅","顺","来","欢","悦","丰","隆","万","腾","丽","仁","朝代","豪","强","神","仙","珍","荣","靓","超","利","棒","财","师","寿"
        \end{itemize}
    \end{CJK*}
    \item English
        \begin{itemize}
            \item \textbf{Location}:"chinese", "mandarin", "asia", "cathay", "china","Xi’an", "ShanDong", "Hunan", "beijing", "Hong Kong", "shanghai", "tokyo", "canton"
            \item \textbf{Culture}:"Panda", "Dragon", "Phoenix","Great Wall", "Forbidden City", "Golden Gate","Bamboo", "Plum",'lotus','tang','song','han','Jackie Chen','xiaoping'
            \item \textbf{Ambiance}:'house', 'home', 'kitchen', 'dining room',"Brother", "Uncle", "Grandma", "Mama","friend", "friendship","Hometown", "home",'hut', 'village', 'inn', 'bistro','garden','pavillion','palace'
            \item \textbf{Specialty}:"Duck", "Fishball", "Sheep", "Seafood", "Tofu", "BBQ", "Buffet", "Green Tea", "Boba Tea", "Tea", "Dim Sum", "Yum Cha", "Dumpling", "Bao", "Egg Roll", "Hot Pot", "Pot", "Noodle", "Rice", "Noodles", "Congee", "Porridge", "Halal", "Veggie", "Vegan", "Kosher", "Wok", "Rice", "Snacks"
            \item \textbf{Positivity}:'best', 'great', 'perfect', 'elite', 'legendary', 'No.1', 'OK', 'super','Delicious', 'Tasty', 'Yummy', 'Luscious', 'Gourmet', 'Fresh','gold', 'golden', 'silver', 'diamond', 'jade','specialist', 'master', 'artisan','8', 'lucky', 'fortune','Happy', 'Joy', 'Sunny', 'Pleasure', 'Nice','Grand', 'Feast', 'Jumbo','King', 'Empress', 'Empire', 'Regent', 'Royal', 'VIP', 'Dynasty','red'
            \item \textbf{Creativity}:'with','to','and','in','out'
        \end{itemize}
\end{itemize}
For the categories that can not be captured through keywords, we devised different methods for each. The Chinese Personal\_Name category is detected using a Chinese NER model that identifies the names of people in Mandarin. The Chinese Creativity category is detected through comparing a name with well-known Chinese proverbs, and a name is considered creative if its similarity score given by the model is above 0.8, which means that it resembles a Chinese proverb. The English Personal\_Name category is detected through a NER model as well, and the English Romanization category is detected using landid, a language identification tool.

\subsection{LLM-based Method}
\label{appendix:llm_model}

Here we provide an example of the name frame \textbf{Ambiance}:
\begin{itemize}
    \begin{CJK*}{UTF8}{gbsn}
        \item Chinese (zero-shot)\par
        你正在为美国中餐馆名称研究数据集进行二元名称框架编码。\par
        请使用谨慎、稳定的研究编码判断。你可以在心里核对规则，但最终答案必须只包含一个数字。\par
        任务：判断该餐厅名称是否属于“氛围”类别。\par
        如果属于，返回1；如果不属于，返回0。\par
        最终回答必须正好是一个字符：0或1。\par
        餐厅名称：\textnormal{\{ CN\_NAME \}}\par
        “氛围”的定义：氛围指名称唤起就餐环境、舒适感、待客感、家庭感、友情、家的感觉、乡愁、悠闲雅致、热闹感或适合用餐的邀请感。\par
        只要名称中任一部分明确符合该类别，整个名称就编码为1。\par
        只使用定义进行判断；不要依赖示例，因为此提示不提供示例。\par
        只回答0或1。\par
        
        \item Chinese (few-shot)\par
        你正在为美国中餐馆名称研究数据集进行二元名称框架编码。请尽量复现论文中的人工标注标准。\par
        任务：判断该餐厅中文名是否属于“氛围”类别。\par

        餐厅名称：\textnormal{\{ CN\_NAME \}}\par
        
        最终回答必须只包含一个字符：0或1。\par
        如果属于“氛围”，回答1；如果不属于，回答0。\par
        
        研究编码规则：\par
        标为1：名称中出现明确的餐厅空间感、就餐环境、亲切感、家庭感、友情、乡愁、雅致感、热闹感或传统店铺意象。\par
        标为0：名称只是地名、菜系、菜品、文化符号、人名/姓氏、普通品牌音译，或只是吉祥积极词但没有空间/氛围含义。\par
        
        稳定标为1的线索：\par
        - 店铺空间：楼、酒楼、酒家、阁、亭、轩、堂、居、府、坊、馆、小馆、屋、厨、厨房。\par
        - 环境场景：园、苑、宫、庭、村、乡、庄、湾、城、巷、吧、茶座。\par
        - 亲切氛围：家、老家、人家、故乡、朋友、相聚、团圆、暖、雅、闲。\par
        
        稳定标为0的线索：\par
        - 只有普通业态词：餐馆、餐厅、饭店，如果没有其他氛围线索，标0。\par
        - 纯菜品或烹饪方式：面、饺子、火锅、烧烤、香锅、外卖、快餐。\par
        - 只有地名、文化符号或积极词：北京、四川、龙、熊猫、福、金、旺。\par
        
        以下线索和例子并不穷尽；请把规则应用到完整餐厅名。\par
        
        正例：\par
        - 松月小馆：1\par
        - 竹里雅轩：1\par
        - 云灯食坊：1\par
        - 暖家厨房：1\par
        
        反例：\par
        - 川椒面点：0\par
        - 金福龙：0\par
        - 台北牛肉面：0\par
        - 王师傅水饺：0\par
        
        请按以上规则判断，只回答0或1。\par
    \end{CJK*}
    
\item Engilish (zero-shot)\par
You are coding Chinese restaurant names in the United States for a binary name-frame dataset.\par
Apply the coding rubric consistently. Think through edge cases silently, then return a single-token answer: 0 or 1.\par
Task: decide whether the restaurant name belongs to the Ambiance category.\par
Return 1 if yes. Return 0 if no.\par
The final response must be exactly one character: 0 or 1.\par
Restaurant name: \{EN\_NAME\}\par
Definition of Ambiance: Ambiance refers to words that evoke dining atmosphere, comfort, hospitality, family, friendship, home, nostalgia, leisurely refinement, or an inviting place to eat.\par
If any part of the name clearly satisfies the category, code the entire name as 1.\par
Use the definition only; do not rely on examples because none are provided.\par
Answer only 0 or 1.\par

\item English (few-shot)\par
You are coding U.S. Chinese restaurant names for a research dataset.\par
Task: decide whether the English restaurant name belongs to the Ambiance category.\par
Restaurant name: \{EN\_NAME\}\par

Return exactly one character: 1 or 0.
Rule: mark 1 when the name evokes a dining place, atmosphere, hospitality, family warmth, friendship, home, nostalgia, leisure, refinement, comfort, or an inviting place to eat.\par

High-confidence 1 cues:\par
- inviting place words: garden, house, pavilion, villa, lounge, hall, inn\par
- warmth or hospitality words: home, family, friendship, cozy, hearth, welcome\par
- atmosphere or leisure words: moonlit, breeze, delight, elegant, peaceful, gathering\par
- family member words: Mama, Uncle, Grandma, Auntie\par

High-confidence 0 cues:\par
- cuisine or geography alone: Chinese, Asian, Beijing, Shanghai, Sichuan, Hunan\par
- food or service format alone: noodle, dumpling, hot pot, buffet, express, takeout\par
- positive/status words without atmosphere: lucky, golden, fortune, best, royal\par

The cues and examples given here are not exhaustive; apply the rule to the full restaurant name.\par

Independent examples:\par
- Beijing Bistro: 1\par
- Grandma's Kitchen: 1\par
- Peony Pavilion: 1\par
- Mama Noodle: 1\par
- Fengtian Bistro \& Lounge: 1\par
- Tianjin Pepper Noodle: 0\par
- Chef Fong Express: 0\par
- Auspicious Gold Wok: 0\par
- Pacific Soy Buffet: 0\par

Answer only 0 or 1.

\end{itemize}

\subsection{Full Model Performance}
\label{appendix:full_nfd_performance}

The Table \ref{tab:nfd_performance} is an aggregated version of Table \ref{tab:nfd_performance_all} below. Table \ref{tab:nfd_performance_all} shows the detailed performance of different LLM models with three prompting strategies. 

\setlength{\extrarowheight}{3pt}
\begin{table*}[ht!]
    \centering
    \footnotesize
    \resizebox{0.99\textwidth}{!}{%
\begin{tabular}{p{3.5cm}cccccccc}
    {\textbf{Model}} & {\textbf{Personal Name}} & {\textbf{Speciality}} & {\textbf{Positivity}} & {\textbf{Culture}} & {\textbf{Location}} & {\textbf{Ambiance}} & {\textbf{Creativity}} & {\textbf{Romanization}} \\
    \Xhline{3\arrayrulewidth}
    \multicolumn{9}{c}{\textit{\textbf{Chinese Names} (Weighted F1)}} \\
    Population Avg. & $94.93$ & $95.45$ & $88.13$ & $81.07$ & $91.50$ & $86.65$ & $96.30$ & $-$ \\
    Expert Rules & $88.09$ & $97.39$ & $90.81$ & $94.79$ & $88.18$ & $93.44$ & $95.20$ & $-$ \\
    Qwen3.7-Max (zero-shot) & $95.14$ & $97.55$ & $86.49$ & $64.45$ & $88.25$ & $77.50$ & $96.89$ & $-$ \\
    Qwen3.7-Max (few-shot) & $94.99$ & $\mathbf{98.86}$ & $92.38$ & $94.15$ & $92.80$ & $\mathbf{96.08}$ & $96.46$ & $-$ \\
    Kimi K2.6 (zero-shot) & $94.06$ & $94.29$ & $81.48$ & $62.38$ & $87.82$ & $73.86$ & $96.66$ & $-$ \\
    Kimi K2.6 (few-shot) & $94.23$ & $97.42$ & $88.64$ & $94.93$ & $94.37$ & $93.46$ & $\mathbf{97.05}$ & $-$ \\
    GPT-5.5 (zero-shot) & $95.44$ & $85.20$ & $88.17$ & $55.74$ & $90.35$ & $83.83$ & $92.85$ & $-$ \\
    GPT-5.5 (few-shot) & $\mathbf{96.01}$ & $98.02$ & $\mathbf{92.68}$ & $95.38$ & $93.99$ & $94.71$ & $96.88$ & $-$ \\
    Claude Sonnet 4.6 (zero-shot) & $94.73$ & $94.49$ & $85.09$ & $84.85$ & $89.21$ & $78.82$ & $96.76$ & $-$ \\
    Claude Sonnet 4.6 (few-shot) & $94.83$ & $97.75$ & $90.13$ & $\mathbf{96.66}$ & $\mathbf{95.20}$ & $94.95$ & $96.88$ & $-$ \\
    \Xhline{3\arrayrulewidth}
    \multicolumn{9}{c}{\textit{\textbf{English Names} (Weighted F1)}} \\
    Population Avg. & $95.32$ & $93.96$ & $89.83$ & $91.04$ & $83.30$ & $82.64$ & $97.22$ & $87.65$ \\
    Expert Rules & $85.83$ & $85.83$ & $82.46$ & $93.95$ & $92.02$ & $\mathbf{91.43}$ & $74.22$ & $72.42$ \\
    Qwen3.7-Max (zero-shot) & $91.19$ & $96.52$ & $88.72$ & $84.86$ & $77.28$ & $72.83$ & $98.27$ & $78.55$ \\
    Qwen3.7-Max (few-shot) & $96.79$ & $\mathbf{97.46}$ & $91.61$ & $96.95$ & $91.11$ & $90.34$ & $\mathbf{99.21}$ & $91.79$ \\
    Kimi K2.6 (zero-shot) & $92.95$ & $88.59$ & $89.04$ & $86.66$ & $68.97$ & $66.75$ & $97.27$ & $85.11$ \\
    Kimi K2.6 (few-shot) & $96.27$ & $96.72$ & $89.41$ & $\mathbf{97.64}$ & $91.61$ & $88.44$ & $97.60$ & $90.57$ \\
    GPT-5.5 (zero-shot) & $93.49$ & $89.36$ & $86.63$ & $83.54$ & $78.26$ & $87.04$ & $94.67$ & $83.19$ \\
    GPT-5.5 (few-shot) & $\mathbf{98.45}$ & $97.27$ & $\mathbf{93.14}$ & $97.49$ & $\mathbf{95.18}$ & $84.33$ & $97.01$ & $93.12$ \\
    Claude Sonnet 4.6 (zero-shot) & $95.80$ & $88.46$ & $87.62$ & $84.84$ & $77.56$ & $85.42$ & $95.74$ & $84.78$ \\
    Claude Sonnet 4.6 (few-shot) & $97.62$ & $97.27$ & $92.46$ & $96.35$ & $86.46$ & $85.95$ & $97.98$ & $\mathbf{94.08}$ \\
    \Xhline{3\arrayrulewidth}
\end{tabular}
    
}
    \caption{Performance of different approaches on Name Frame Detection in both English and Chinese ACR names. ``Expert Rules'' represents the rule-based system based on lexicons and heuristics. We report performance for Qwen3.7-Max, Kimi K2.6, GPT-5.5, and Claude Sonnet 4.6 under two prompt settings: (1) ``zero-shot'', (2) ``few-shot'', respectively, as described in Section \ref{sec:nfd_task}.}
        \label{tab:nfd_performance_all}
\end{table*}

\subsection{GPT Translation Guideline}
\label{appendix:gpt_translation_guideline}


Here is a list of rules to translate Chinese restaurant names to English:\par
    1. Translate each name character by character. No need to care about English grammar. Some examples:\par
        \begin{CJK*}{UTF8}{gbsn}
        '西海楼' 'West Sea House'\par
        '福满楼' 'Luck Full House'\par
        '锦灿' 'Golden Splendor'\par
        '二舅烤肉' 'Second Uncle Barbecue'\par
        \end{CJK*}
    2. If a character or a phrase has a Chinese meaning and doesn’t look like a person’s name, return the literal translation of this character or phrase instead of the Romanized version and please pay special attention to this.
    
    3. Make sure there’s as little Romanization in the translations as possible. Here are some examples for your reference. For instance,\par
    \begin{CJK*}{UTF8}{gbsn}
        '明苑阁' 'Bright Garden Pavillion'\par
        '川香园' 'Sichuan Fragrant Garden'\par
        '那家老院子' 'Na's Old Yard'\par
        '芳芳小食' 'Fang Fang Snacks'\par
    \end{CJK*}
Make sure to refer to these examples when you do the translation.\par
Give me the literal english translation of the Chinese restaurant name \{CN\_NAME\}. Only return the english translations and don't return the Chinese names. No explanation needed. No double or single quotes needed.

\subsection{Translation Model Selection}
\label{appendix:gpt_translation_performance}

We compared text similarity scores between GPT-5.5, Claude Snonnet 4.6 translations, and Google Translate. Both AI models outperformed Google Translate. While both AI models achieved comparable results, Claude 4.6 Sonnet achieved marginally higher scores in five out of nine metrics compared to GPT-5.5. Therefore, we decided to use Claude Snonnet 4.6 to translate the Chinese names of our sampled restaurants.
\setlength{\extrarowheight}{3pt}
\begin{table*}[ht!]
    \centering
    \footnotesize
    \resizebox{0.7\textwidth}{!}{%
    \begin{tabular}{lrrr}
        \Xhline{3\arrayrulewidth}
        & \multicolumn{1}{c}{\textbf{Google Translate}} &
        \multicolumn{1}{c}{\textbf{GPT-5.5}} &
        \multicolumn{1}{c}{\textbf{Claude Sonnet 4.6}} \\
        \Xhline{3\arrayrulewidth}
        Cosine & $0.6550$ & $0.7598$ & $\mathbf{0.7706}$ \\
        BERT F1 & $0.9198$ & $0.9404$ & $\mathbf{0.9432}$ \\
        BERT Precision & $0.9161$ & $0.9417$ & $\mathbf{0.9434}$ \\
        BERT Recall & $0.9241$ & $0.9392$ & $\mathbf{0.9431}$ \\
        ROUGE-1 & $0.4722$ & $\mathbf{0.5975}$ & $0.5920$ \\
        ROUGE-2 & $0.3003$ & $0.3692$ & $\mathbf{0.3781}$ \\
        ROUGE-L & $0.4628$ & $\mathbf{0.5914}$ & $0.5829$ \\
        ROUGE-Lsum & $0.4628$ & $\mathbf{0.5914}$ & $0.5829$ \\
        BLEU & $0.0220$ & $\mathbf{0.0488}$ & $0.0424$ \\
        \Xhline{3\arrayrulewidth}
\end{tabular}}
    \caption{Translation quality comparison across Google Translate, GPT-5.5, and Claude Sonnet 4.6.}
    \label{tab:translation_performance}
\end{table*}


\section{Interview Details}


\subsection{Interview Procedure}
\label{app:interviewProcedure}
We conducted semi-structured interviews with ten ACR owners. 
We refer to the participants from R1 to R10. 
Details for each respondent can be found in Table~\ref{tab:interview_table}. We follow the guideline\footnote{\url{https://www.mass.gov/doc/massachusetts-rural-towns-list/download}} from Massachusetts State Office of Rural Health to classify the areas. To attain a diverse sample, we employed a stratified sampling approach to ensure that we reach owners from both urban and rural areas.
We also sampled restaurants based on whether their Mandarin and English names were a close match or a mismatch.
R9 was the only restaurant that didn't have a Mandarin name.

We reached out to the selected restaurants through walk-ins or calls. 
Each interview was held in Mandarin and lasted around 30 minutes to an hour.
We asked owners how the Mandarin and English restaurant names were created. 
We also asked questions about personalizing the names to different customer groups, how the restaurant location impacted the naming process, and how the restaurant reflects aspects of the owner's heritage and identity.
Appendix~\ref{app:interviewGuide} contains our interview guide. 

Since a majority of participants didn't feel comfortable with recording the conversations, we had a designated note-taker record detailed notes in English, which we used to analyze the interviews. 
Three authors went through the notes and coded them for themes using affinity diagramming, a popular analysis method in user interface design~\cite{lucero2015using} where text data is segmented into notes and then clustered into groups.
Through multiple iterations, along with periodic discussions with
the rest of the research team, we identified several key themes that impact restaurant naming decisions. 

\setlength{\extrarowheight}{3pt}
\begin{table*}[t!]
    \centering
    \scriptsize
    \begin{tabular}{ccccccc}
        {\textbf{Respondent}}  &
        {\textbf{Gender}} &
        {\textbf{English Proficiency}}  &
        {\textbf{Customer Base}} &
        {\textbf{Area}} &
        {\textbf{Opening Year}} &
        {\textbf{Name Match}} \\
         \Xhline{3\arrayrulewidth}
        R1 & Female & Proficient & Weekdays: 20\% Chinese; Weekends: 50\% & Urban & 2017 & Close match \\
        R2 & Female & Advanced & 50\% Chinese & Urban & 2011 & Mismatch\\
        R3 & Female & Intermediate & 45\% Chinese & Urban & 2000 & Mismatch \\
        R4 & Female & Intermediate & Weekdays: 50\% Chinese; Weekends: 75\% & Urban & 2003 & Close match \\
        R5 & Male & Basic & 80\% Chinese & Urban & 1993 & Mismatch \\
        R6 & Male & Proficient & 40\% Chinese & Urban & 2010 & Mismatch \\
        R7 & Male & Fluent & 50\% Chinese & Urban & 2016 & Mismatch \\
        R8 & Male & Proficient & 40\% Chinese & Urban & 2005 & Close match \\
        R9 & Male & Intermediate & 1\% Chinese & Rural & 2012 & No Chinese name\\
        R10 & Male & Intermediate & 1\% Chinese & Rural & 2000 & Close match\\
        \Xhline{3\arrayrulewidth}
    \end{tabular}
    \caption{Information about 10 ACR respondents that we interviewed. Each respondent is an owner, co-owner, or founder of an ACR in Massachusetts. The gender, English proficiency, percentage of Chinese customers, whether the restaurant is located in an urban or rural area, and the closeness between the Mandarin and English names are reported.}
        \label{tab:interview_table}
\end{table*}

\includepdf[pages=1,scale=1,pagecommand=\subsection{Interview Guide}\label{app:interviewGuide}]{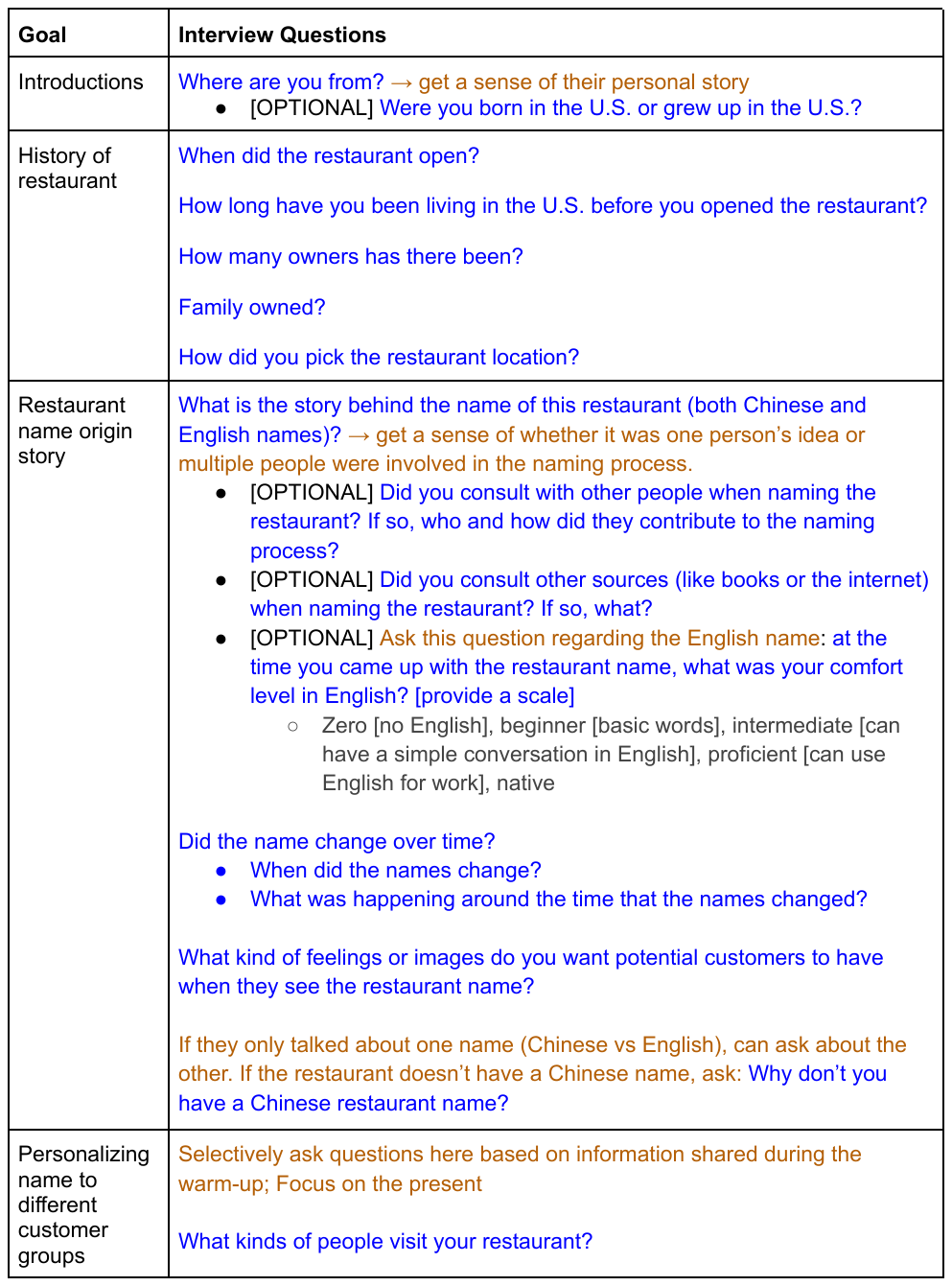}
\includepdf[pages=2,pagecommand={}]{fig/cleaned_interview_plan.pdf}

\subsection{Expanded Interview Results}
\label{app:interviewResults}

\subsubsection{Impact of Customer Group}

All owners strategically think about their customer groups, especially in regards to age and cultural background, observing how their customers change over time. For example, R10 caters their food to the local rural population, which is less than 1\% Chinese, while R5 caters their food to the international Chinese students who come to Chinatown and want to try new foods. 

In many cases, the owners’ awareness of their customer base directly impacted their choice in Chinese and English restaurant names. ACR Mandarin names help attract Chinese-speaking customers. R2 chose their Chinese name to attract Chinese customers who enjoy certain regional dishes, which is made clear in the name. Further, R2 is well-aware that the mismatch in their Chinese and English names denotes two different regions of China, which sparks the curiosity of many of their Chinese customers; \emph{``Chinese customers…want to investigate and try it.''} About \emph{``8-9\% of young customers ask where the name is coming from.''} All but two ACRs came up with their Chinese names first. R1 noted that “Chinese names are easy to understand, because the \emph{``language is very rich,''} explaining that their particular Chinese name holds a double meaning. 

Similarly, ACR English restaurant names attract largely American customers. In the case of R9, which only has an English name, there is \emph{``no need''} for a Chinese name, as it is located in a rural area where \emph{``there are no Chinese customers.''} Many ACR owners chose their English names because they are familiar and easy to understand for American customers. R2 decided upon their English name first, because the \emph{``name tells [American customers] it’s an Asian food.''} R6, whose Chinese and English names are mismatches, said of their English name: \emph{``We want to accommodate for American taste and have to change to fit their palette.''} R4 named their restaurant after a popular movie featuring Chinese characters, saying \emph{``it is good to have a simple name to be more acceptable by customers.''} Some ACR owners caught onto Western trends of trying new cuisines, and named their restaurants in accordance with the trends. R6’s name brings up associations with a popular Chinese dish at a time where journalists were reporting on it, which \emph{``made American customers curious.''} R8, whose two names are a close match, expects American customers to google the restaurant name and \emph{``be curious about the place.''} And for some ACR owners like R1 and R8, the English name plays a more active role, acting as a means to correct misconceptions of Chinese restaurants in America as \emph{``cheap, not very good, not clean''} (R1). R1’s goal was to “attract westerners, have them try the Chinese food and change their mind,” establishing R1 as a \emph{``social destination, not just place to eat.''} Accordingly, \emph{``Westerners can say the name” and “it sounds sweet''} (R1). 

\subsubsection{Impact of Location}

Several ACRs highlighted the importance of opening a restaurant in a \emph{``good neighborhood''} with \emph{``better school districts''} (R2).
R3 chose to open her restaurant near a prestigious university to attract students, researchers, and professors.
Those universities have a large group of Chinese international students who regularly eat at Chinese restaurants, motivating the need for a Mandarin name. 
R1 opened her restaurant in an area with a lot of development, which happens to be close to her former workplace.
She wanted to \emph{``create a place for former co-workers to come and keep in touch''}, which was one reason why she incorporated her personal name in both the Mandarin and English restaurant names. 

We observed a different relationship between location and naming decisions in Boston's Chinatown, which has the largest concentration of ACRs in Massachusetts. 
R7 shared that a restaurant in Chinatown has to be \emph{``unique to stand out so that they can have business. There is a lot of competition for the same type of food, and other restaurants have been open for longer.''}
R7 identified a type of Chinese food that was not being sold in Chinatown. He highlighted aspects of the food in both the Mandarin and English names of his restaurant to stand out from other options in the area. 
Chinatown is drastically different from rural areas, where there are very few ACRs. 
R9 decided to open a restaurant in a rural town to \emph{``escape Chinatown.''}
He and his partner picked their current location because it is \emph{``adventurous''}, and they wanted to \emph{``start a new life with more freedom.''}
The newfound freedom is reflected in R9's choice of naming the restaurant in memory of his daughter and deciding not to create a Mandarin name.

\subsubsection{Impact of Personal Experience}

A few ACR owners shared how their personal experience, including their family, origin, and personality, influenced their choice of restaurant names. Both R3 and R9 included the names of their daughters to ``\textit{show love}''. R1 included her Chinese given name in the ACR name to memorize her dad who was ``\textit{an artist who came up with the name}'' (R1). R1's dad passed away in China while she was still studying in America and did not make it to see him as the only daughter. A few owners (R1, R2, R4) also mark their or their partners' origin in the ACR names such as ``\textit{Hunan}'', ``\textit{Taiwanese}''. R1 shared that she did this to ``\textit{introduce my hometown to this area}'' and expressed, ``\textit {I miss my family in the hometown}.'' 6 out of 10 respondents are from Fuzhou City, which is a community has long been known for its significant presence in the Chinese restaurant industry in the United States. It is interesting to see that none of the six ACRs owned by Fuzhounese encode their origins in the names. R8, an active member from the Fuzhouese community, confirmed our observation that most ACRs in the state are owned by immigrants from Fuzhou City, Fujian Providence. At last, we found that some owners wanted the name to ``\textit{be personal}'' (R1) and ``\textit{become a form of self expression}'' (R7). R7, who decorated and named his ACR in a unique and fashionable manner, shared that he ``\textit{did not care if other people like the name}'' when he created and copyrighted the name. However, some owners such as R3 had a different opinion, sharing that she did not like ``\textit{her identity to be encoded in the restaurant names}.''

\section{Prolific Setup}

We recruited Prolific workers from around the world whose first language is Chinese with an approval rate between 99\% and 100\%. We divided our sample of 1956 data into 70 Qualtrics surveys, with each survey completed by two prolific workers. We began our survey with a series of instructions and asked that the workers answer in simplified Chinese characters. Then, the workers find the names of restaurants using the links we provided in the survey. Each survey contains 30 - 50 restaurants. Screenshots of the questions can be found below. All our questions were written in both simplified Chinese and English.

\begin{figure}[h]
\includegraphics[width=\columnwidth]{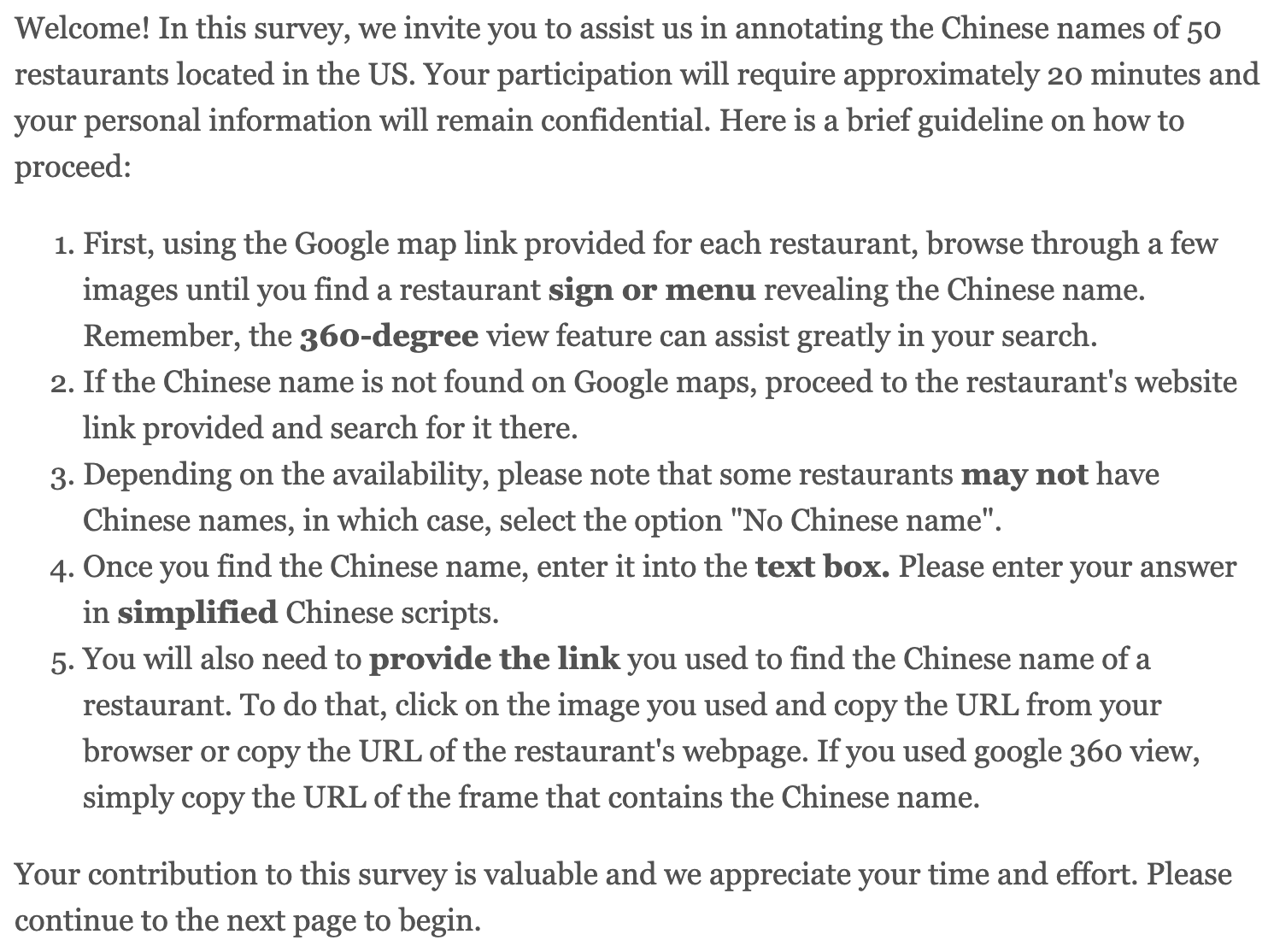}
\caption{Qualtrics survey English instructions }
\label{fig:frame_dist_1}
\end{figure}

\begin{figure}[h]
\includegraphics[width=\columnwidth]{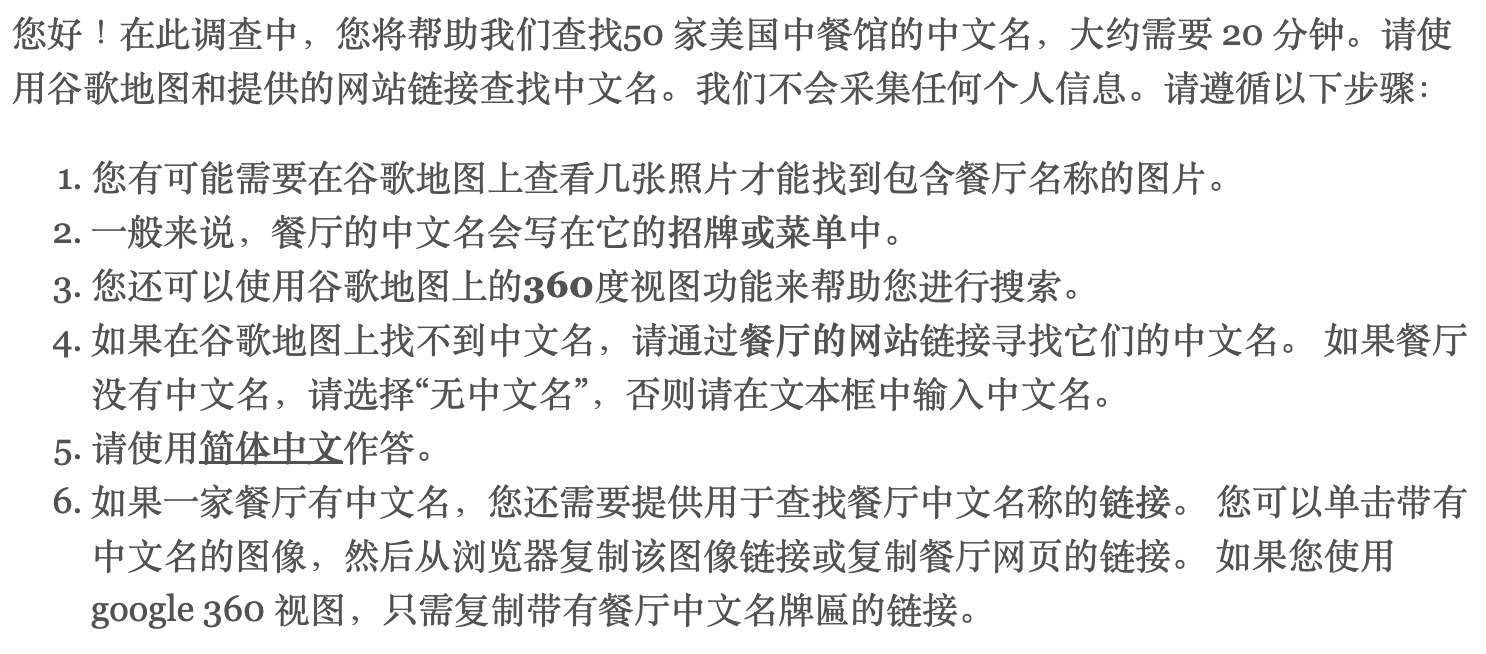}
\caption{Qualtrics survey Chinese instructions }
\label{fig:frame_dist_2}
\end{figure}

\begin{figure}[h]
\includegraphics[width=\columnwidth]{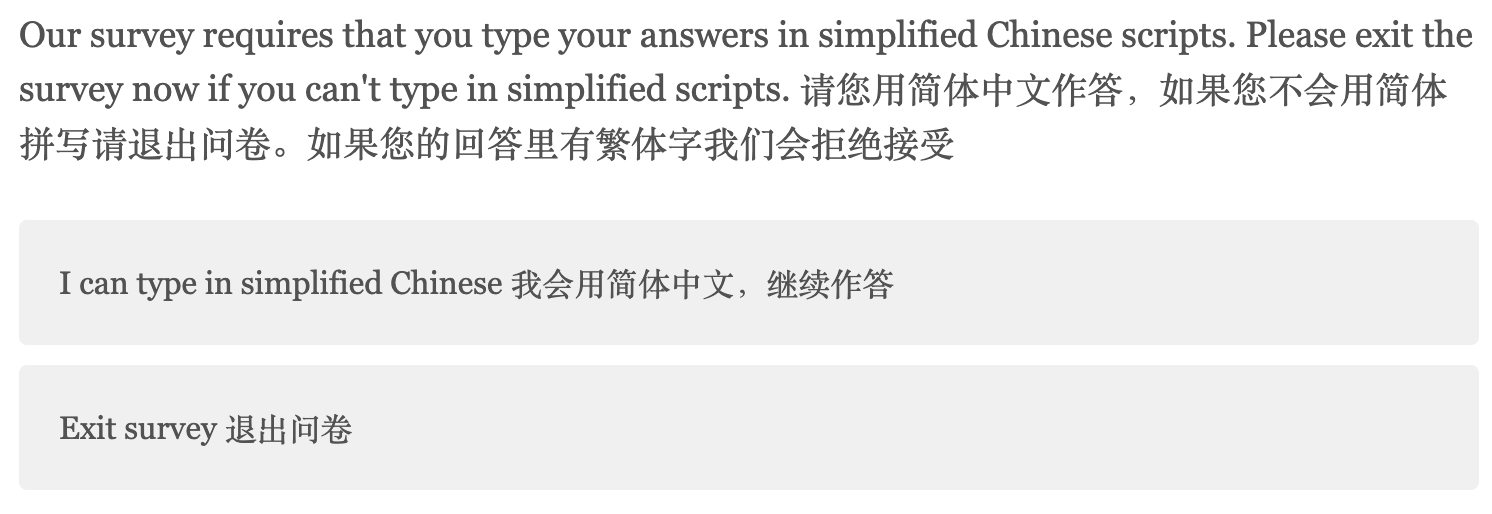}
\caption{Qualtrics survey response instructions }
\label{fig:frame_dist_3}
\end{figure}

\begin{figure}[h]
\includegraphics[width=\columnwidth]{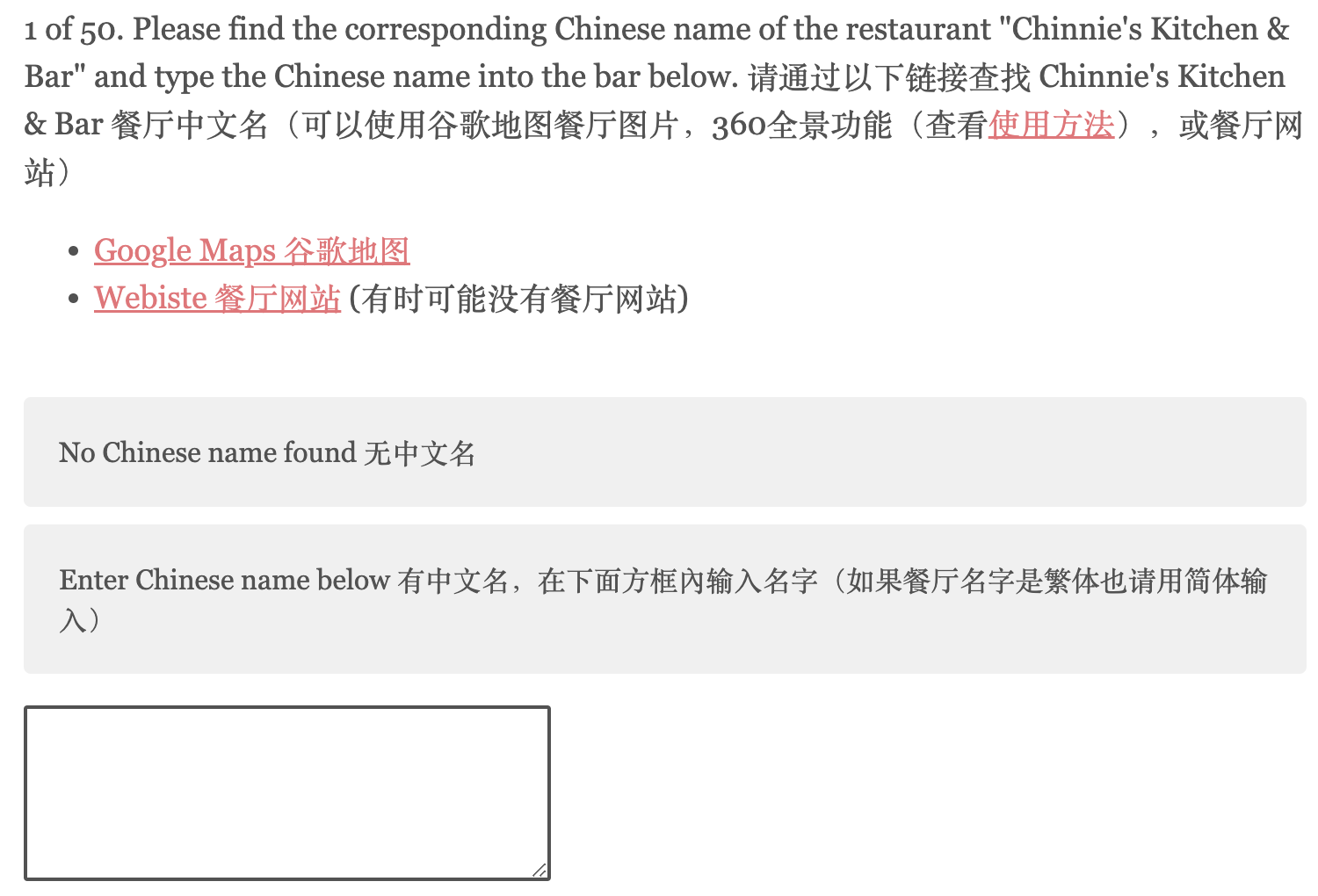}
\caption{Qualtrics survey sample question }
\label{fig:frame_dist_4}
\end{figure}

\subsection{Prolific Worker Demographics}
There are 20 unique participants who took the surveys. They are based in the UK, Canada, New Zealand, and Germany. 12 of them were born in mainland China, 4 in Hong Kong, 2 in the UK, 1 in Malaysia, and 1 in Taiwan. The distribution of their age and sex can be found below. 

\begin{figure}[htbp]
    \centering
    \begin{subfigure}[b]{0.48\textwidth}
        \centering
        \includegraphics[width=\textwidth]{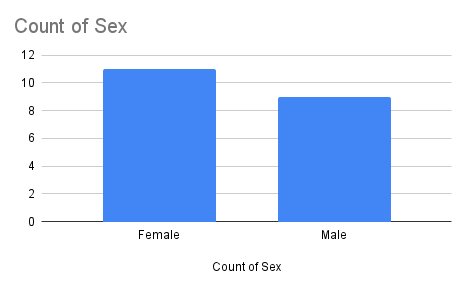} 
        \caption{Prolific workers sex distribution}
        \label{fig:sex_dist} 
    \end{subfigure}
    \hfill 
    \begin{subfigure}[b]{0.48\textwidth}
        \centering
        \includegraphics[width=\textwidth]{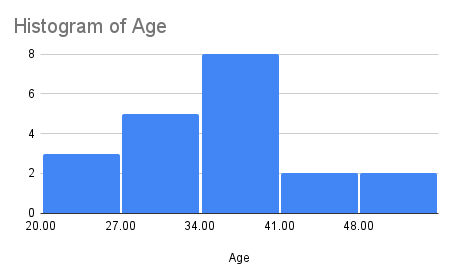}
        \caption{Prolific workers age distribution}
        \label{fig:age_dist} 
    \end{subfigure}
    
    \caption{Demographic characteristics of Prolific workers} 
    \label{fig:demographics}
\end{figure}

\clearpage
\section{Regression Analysis}
\label{appendix:continuous-duration}


\subsection{Multivariate regression}
\label{appendix:regression_analysis_1}
\begin{table}[htbp]
\centering
\caption{Multi-factor regression results}
\label{tab:regression_results}

\begin{threeparttable}
\begin{tabular}{lcccc}
\toprule
 & \multicolumn{2}{c}{Jaccard Distance} & \multicolumn{2}{c}{Cosine Distance} \\
\cmidrule(lr){2-3} \cmidrule(lr){4-5}
Predictor & $\beta$ (SE) & $p$ & $\beta$ (SE) & $p$ \\
\midrule
Intercept
  & 0.5706 (0.010) & 0.000 & 0.5165 (0.007) & 0.000 \\
Chinese population \% (scaled)
  & 0.0128 (0.010) & 0.199
  & -0.0085 (0.007) & 0.195 \\
Time since opening (scaled)
  & -0.0128 (0.010) & 0.194
  & -0.0179 (0.007) & 0.006 \\
Location (rural) 
  & -0.1682 (0.069) & 0.014
  & 0.0181 (0.045) & 0.688 \\
Location (Suburb) 
  & -0.0388  (0.039) & 0.323
  & -0.0360  (0.026) & 0.165 \\

\midrule
$R^2$        & \multicolumn{2}{c}{0.008} & \multicolumn{2}{c}{0.007} \\
Adj. $R^2$   & \multicolumn{2}{c}{0.005} & \multicolumn{2}{c}{0.004} \\
$F$-statistic & \multicolumn{2}{c}{2.989 ($p$ = 0.0180)} & \multicolumn{2}{c}{2.611 ($p$ = 0.0340)} \\
Max VIF & \multicolumn{4}{c}{1.04} \\
\bottomrule
\end{tabular}

\begin{tablenotes}
\small
\item Note: Continuous predictors were standardized (mean = 0, SD = 1)
before fitting. Coefficients therefore reflect the expected change in
the outcome per one standard deviation increase in the predictor.
For location factor, urban category is chosen as reference category as it is the dominate category (n = 1,346) compared to suburb (n = 100) and rural (n = 31). 

\end{tablenotes}

\end{threeparttable}

\end{table}


\subsection{Random Forest regressor}

\label{appendix:regression_analysis_2}

\begin{table}[htbp]
\centering
\caption{Random Forest regression results.}
\label{tab:rf_gridsearch_results}
\begin{threeparttable}

\begin{tabular}{lcc}
\toprule
DV & CV $R^2$ & Train / Test $R^2$ \\
\midrule
Jaccard distance
  & $-0.0782$ & $-0.0164$ / $-0.0131$ \\
Cosine distance
  & $-0.0040$ & $-0.0294$ / $-0.0761$ \\
\bottomrule
\end{tabular}

\begin{tablenotes}
\small
\item Note:  To address class imbalance (Rural $n=31$, Suburb $n=100$, Urban $n=1{,}346$), we used a stratified train-test split and assigned higher sample weights to minority location categories. Hyperparameters were selected via cross-validation to maximize $R^2$, for both distance measures we used \texttt{max\_depth = 3} and \texttt{min\_samples\_leaf = 50}.
\end{tablenotes}
\end{threeparttable}
\end{table}

\end{document}